\documentclass[11pt]{article}
\ifdefined\pdfobjcompresslevel\pdfobjcompresslevel=0\fi
\usepackage[preprint]{acl}
\usepackage[T1]{fontenc}
\usepackage{times}
\usepackage{latexsym}

\usepackage{microtype}
\usepackage{stfloats}
\usepackage{graphicx}
\usepackage{subfig}
\usepackage{amsmath,amssymb,bbm,mathtools}
\usepackage{bm}
\usepackage{booktabs}
\usepackage{threeparttable}
\usepackage{enumitem}
\usepackage{stmaryrd}
\usepackage{breqn}
\usepackage{mdwlist}
\usepackage{pifont}
\usepackage{array}
\usepackage{colortbl}
\usepackage{tikz}
\usetikzlibrary{calc,fit,positioning}
\usepackage{tcolorbox}
\tcbuselibrary{skins,breakable}
\usepackage{listings}
\lstdefinestyle{promptstyle}{%
  basicstyle=\ttfamily\footnotesize,
  breaklines=true,
  breakatwhitespace=false,
  columns=fullflexible,
  keepspaces=true,
  showstringspaces=false,
  upquote=true,
  frame=single,
  framerule=0.3pt,
  framesep=3pt,
  xleftmargin=3pt,
  aboveskip=4pt,
  belowskip=6pt
}
\lstnewenvironment{promptbox}[1][]%
  {\lstset{style=promptstyle,#1}}%
  {}
\newcommand{\promptcaption}[1]{%
  \par\smallskip\noindent\textbf{#1}\par\nobreak\vspace{1pt}%
}

\newcommand{\sysname}{Bazaar}

\mathchardef\mhyphen="2D

\newcommand{\figlab}[1]{\label{fig:#1}}

\tikzstyle{block} = [rounded rectangle, draw,
    minimum width=2.5cm, text centered, minimum height=1cm, node distance=1.5cm]
\tikzstyle{arrow} = [draw, -latex]

\usepackage{tcolorbox}
\tcbuselibrary{skins,breakable}

\newtcolorbox{protobox}[2][]{%
  enhanced,
  title        = {#2},
  attach boxed title to top left={xshift=+3mm,yshift*=-3mm},
  breakable    = false,
  colback      = white, 
  colframe     = black!75,
  fonttitle    = \bfseries,
  colbacktitle = black!10!white,
  coltitle     = black,
  #1
}

\title{Can LLM Agents Price Competitively? A Dynamic Multi-Attribute Auction Benchmark for Agentic Commerce}

\author{Shimaa Ahmed$^{*}$ \quad Yiwei Cai$^{*}$ \quad Mohsen Minaei$^{*}$ \quad Rahul Rachuri$^{*}$ \\
Visa Research}

\begin{document}
\maketitle
\renewcommand{\thefootnote}{*}
\footnotetext{Authors are listed in alphabetical order.}
\renewcommand{\thefootnote}{\arabic{footnote}}

\begin{abstract}
Agentic commerce is moving from concept to deployed infrastructure: payment networks, retailers, and AI platforms are setting the stage for agents to transact on behalf of merchants and consumers. Yet whether the LLMs behind these agents can price competently in real markets, where customer preferences are hidden, competitors adapt in real time, and demand can shift without warning, has not been systematically tested. We introduce \sysname{}, a dynamic sealed-bid benchmark for multi-attribute auction under these conditions. Despite its dynamics, the benchmark is grounded in closed-form customer utilities, enabling exact evaluation. Across 11 frontier LLMs from four providers, the leading agents on customer acquisition (e.g. Gemini~3.1~Pro) are often not the leading agents on profit (e.g. Opus~4.6). The ranking shifts again under demand shocks: agents that learned fastest pre-shock are typically the slowest to revise their beliefs afterwards, while Gemini~3.1~Pro recovers fastest despite not leading on profit. However, even the strongest agent captures less than a third of hindsight-optimal profit, suggesting current LLMs are progressing in agentic commerce but leave substantial headroom.
\end{abstract}

\section{Introduction}
\label{sec:intro}

Language models are beginning to act as economic agents. In commerce, an
agent can negotiate with customers, place bids, choose product bundles, and set
prices on behalf of a buyer or seller
\cite{magnetic25,zhu2025automatedrisky,yu2026shoppingcompanion}. Recent systems
such as Anthropic's Project Vend~\cite{anthropic2025projectvend2}, Microsoft's Magentic Marketplace~\cite{magnetic25}, and
Alibaba's Shopping Companion~\cite{yu2026shoppingcompanion} point toward marketplaces populated by LLM
merchants that respond to customers and competitors in real time
\cite{magnetic25,yu2026shoppingcompanion}. This trend raises a concrete
evaluation question: \emph{when an LLM is placed inside a competitive market
loop, can it learn what customers value, choose profitable offers, and adapt
when demand changes?}

Existing benchmarks capture important pieces of this problem. E-commerce
benchmarks~\cite{peng2024ecellm,chen2025chineseecomqa} evaluate product
knowledge, intent understanding, and recommendation quality on fixed
instances. Economic and game-theoretic benchmarks
\cite{chen2023aucarena,shah2025syntheticlabs,pmlr-v235-bianchi24a,gametheory24,shapira2024glee}
study auctions, bargaining, and negotiation under stylized rules. Neither
setting combines learning, pricing, and belief revision in the repeated,
nonstationary competition that deployed merchants face.

\begin{figure*}[t]
\centering
\resizebox{0.75\textwidth}{!}{\includegraphics{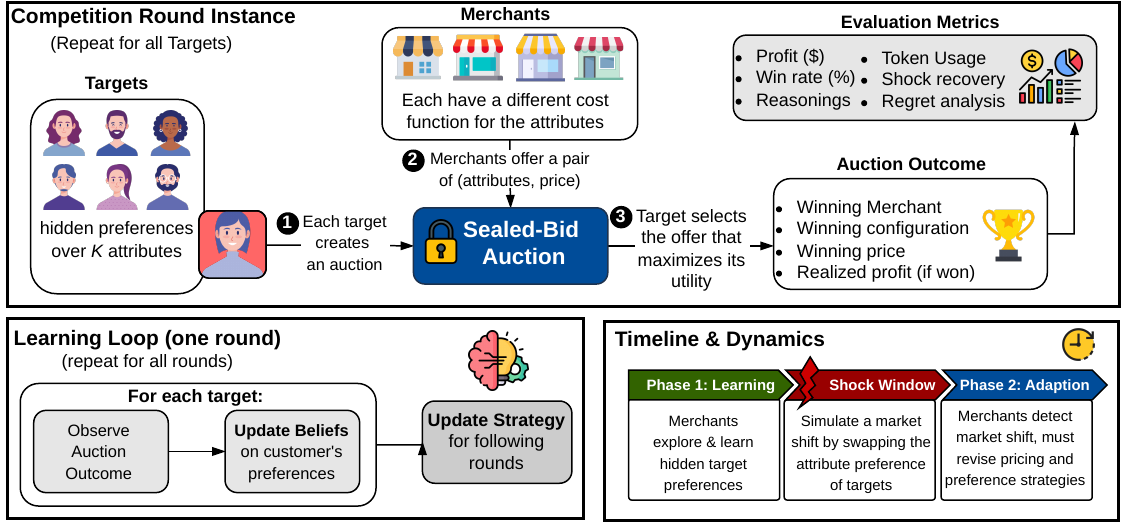}}
\caption{Abstract overview of \sysname{}.}
\label{fig:overview-main}
\end{figure*}

We introduce \sysname{}, a framework for evaluating LLM merchants in
dynamic multi-attribute auction markets. In \sysname{}, an LLM merchant
repeatedly chooses product configurations and prices for customers with
hidden preferences, while competing against rival merchants with different cost
advantages. The agent observes only sparse market feedback, so it must infer
what each customer values, decide how much margin to charge, and update its
strategy over time. Markets also shift: new trends, sudden events, and
seasonal changes can reorder what customers want. To capture this pressure,
\sysname{} introduces unannounced preference shifts midway through the game.
This lets us ask whether agents that learn quickly before a shift can also
revise their beliefs afterward. We evaluate total profit together with whether
an agent chooses the right configuration, charges the right price, and adapts after
demand changes (Figure~\ref{fig:overview-main}). 

This paper makes three contributions:

\vspace{0.5em}
  \noindent\textbf{A diagnostic framework for dynamic LLM pricing
    (\S\ref{sec:model}).} \sysname{} specifies a repeated sealed-bid
    multi-attribute auction protocol with private costs, hidden
    heterogeneous customers, sparse feedback, adaptive competitors, and
    preference shifts. Despite the dynamics, closed-form customer utilities and merchant costs enable an exact evaluation. Its surplus and oracle-regret diagnostics separate performance into preference learning (configuration choice),
    margin extraction (price), and post-shift revision. This lets us
    classify agents into diagnostic failure archetypes (\emph{Loser},
    \emph{Underpricer}) and measure adaptation directly, rather than
    inferring from a single score. Thompson
    Sampling~\cite{thompson1933} and EXP4~\cite{auer2002exp4} provide
    non-LLM baselines for binary win/loss learners in the same action
    space.

  \vspace{0.2em}
  \noindent\textbf{Margin discipline is the discriminating skill, and thinking
    budget moves models along a two-dimensional failure surface
    (\S\ref{sec:results}).} Across 11 frontier LLMs, profit is almost
    perfectly aligned with margin per win ($r{=}0.99$), and less so with
    win rate ($r{=}0.88$). Increasing thinking effort shifts models along
    the failure surface: it lowers loss regret (fewer missed wins) but raises
    underprice regret (leaving money on the table), moving the same base
    model from a \emph{Loser} archetype toward an \emph{Underpricer}. This
    shift is economically large: GPT-5.4 (high) earns $7.3\times$ more than
    GPT-5.4 (none), and Opus~4.6 (adaptive, xhigh) substantially outperforms
    Opus~4.6 (off, high).

  \vspace{0.2em}
  \noindent\textbf{Two distinct paths to the top of the leaderboard, and a strong-learner / weak-adapter reversal under shocks (\S\ref{sec:results-shock}).}
  Opus~4.6 (adaptive, xhigh) leads on profit at \$2{,}976 through margin discipline, earning \$2.22 per win. Gemini~3.1~Pro (dynamic, high) leads on win rate (77.9\%), recovers fastest from preference shocks, writes strategies roughly a third the length of competitors, and is also the most cost-effective agent we evaluate. The shock also reveals a \emph{strong-learner, weak-adapter} pattern:
  the models that climb fastest before the shock (GPT-5.3 and Opus~4.5,
  gaining $+46$ and $+40$~pp in win rate over rounds 1--30) are among the worst at adapting afterward, losing $-19$~pp on average and not
  recovering. A static benchmark would never surface this failure mode.

Our work connects four threads: LLM agents in auctions and negotiation,
multi-agent LLM benchmarks, LLMs in e-commerce, and algorithmic pricing
with regret-based evaluation. Appendix~\ref{app:related} discusses prior
work in each.

\section{\sysname{}: A Dynamic Multi-Attribute Pricing Framework}
\label{sec:model}

In this section, we define a framework for evaluating whether a merchant can
learn profitable pricing policies in repeated, competitive, partially observed
markets. In each round, merchants configure a product and set a price for each
customer. The customer selects the offer that maximizes its utility, and
these utilities may change over time. This abstraction captures a core
challenge of agentic commerce: a merchant must learn what each customer values,
how much margin it can charge, and when earlier beliefs should be revised. We
use \emph{focal merchant} to denote the merchant whose behavior is being
measured; in our experiments this is the LLM agent under evaluation,
but the framework definitions are merchant-agnostic. Figure~\ref{fig:overview}
shows the market schematic.

\begin{table}[t]
\centering
\small
\begin{tabular}{ll}
\toprule
\textbf{Symbol} & \textbf{Meaning} \\
\midrule
$\mathcal{M}$, $f\in\mathcal{M}$ & Merchants; focal merchant \\
$\mathcal{I}$ & Customers \\
$t\in\{1,\ldots,T\}$ & Global round index \\
$K, L$ & Attributes and levels per attribute \\
$\mathbf{x}\in\{1,\ldots,L\}^{K}$ & Attribute-level configuration \\
$p$ & Offered price \\
$V_{i,t}(\mathbf{x})$ & Customer $i$'s hidden value at round $t$ \\
$U_{i,t}(\mathbf{x},p)$ & Customer utility for offer $(\mathbf{x},p)$ \\
$c_m(\mathbf{x})$ & Merchant $m$'s production cost \\
$\pi_m(\mathbf{x},p)$ & Merchant $m$'s profit if it wins \\
\bottomrule
\end{tabular}
\caption{Core notation for the framework. Diagnostic symbols ($q_{i,t}$, $\pi^{\star}_{i,t}$,
$S_{i,m,t}$) are introduced inline in \S\ref{sec:model-surplus}.}
\label{tab:notation}
\end{table}

\subsection{Repeated Multi-Attribute Auctions}
\label{sec:model-mechanics}

Let $\mathcal{M}$ be the set of merchants and $\mathcal{I}$ the set of customers.
Time is divided into global rounds $t\in\{1,\ldots,T\}$. In every round, each
customer $i\in\mathcal{I}$ runs a sealed-bid auction among all merchants.
A bid from merchant $m$ to customer $i$ is
\vspace{-0.3em}
\begin{equation}
  b_{m,i,t}=(\mathbf{x}_{m,i,t},p_{m,i,t}),
\end{equation}
where $\mathbf{x}_{m,i,t}=(x_1,\ldots,x_K)\in\{1,\ldots,L\}^{K}$ is a product
configuration over $K$ attributes with $L$ levels each, and $p_{m,i,t}$ is a
price. Each global round therefore gives the focal merchant one
customer-specific outcome per auction.

\subsection{Customers, Costs, and Profit}
\label{sec:model-costs}

At round $t$, customer $i$ has an additive valuation over configurations,
$V_{i,t}(\mathbf{x}) = \sum_{k=1}^{K} V_{i,t,k}[x_k]$, and the utility of an
offer $(\mathbf{x},p)$ is $U_{i,t}(\mathbf{x},p)=V_{i,t}(\mathbf{x})-p$. We use
additive valuations for simple, interpretable evaluation, though the framework
can support other utility functions. Merchant costs are also additive:
\vspace{-0.5em}
\begin{equation}
  c_m(\mathbf{x}) = c_{m,0} + \sum_{k=1}^{K} C_{m,k}[x_k],
  \label{eq:cost-general}
\end{equation}
where lower $C_{m,k}$ on attribute $k$ encodes specialization. Customer values
and merchant costs are represented as curves over levels, since the value or
cost of moving between adjacent levels need not be linear.
Section~\ref{sec:setup-instantiation} specifies the concrete values of $K$ and
$L$. If merchant $m$ wins with offer $(\mathbf{x},p)$, its profit is
$\pi_m(\mathbf{x},p)=p-c_m(\mathbf{x})$, with bids constrained to satisfy
$p\ge c_m(\mathbf{x})$.

\subsection{Information and Memory}
\label{sec:model-information}

Before bidding, the focal merchant observes its own cost structure, the
customer identity, its stored belief for that customer, its global strategy,
and its financial state. It does not observe customer values $V_{i,t}$,
competitors' costs, or competitors' bids. After a customer-level auction
resolves, all merchants observe the winner's identity, configuration, and
price, while the winner additionally observes realized profit. Losing merchants
do not see the losing bids, the customer utility, or the utility gap to the
winner. A loss therefore conveys only the constraint that the winning offer was
preferred to the focal merchant's offer.

\subsection{Market Shifts}
\label{sec:model-shock}

The components above define a stationary market. With enough interactions, a
successful merchant should be able to form stable beliefs about each customer
and exploit them. Real markets, however, can change because of seasonal demand,
social trends, supply shocks, or external events. We isolate adaptation by
holding merchant costs fixed and changing customer valuations.

A shift is parameterized, for each customer $i$, by a switch round $s_i$ and a
transformation $\mathcal{T}_i$ that maps pre-shift value curves to post-shift
curves.
Crucially, the focal merchant receives no explicit shift indicator. It must
detect change from the same feedback used throughout: winner identities,
winning configurations, and winning prices.

This design makes the shock an adaptation problem rather than a separate
cold-start task. The transformation can preserve part of a customer's previous
structure while changing the mapping from attributes to value, so the focal
merchant must decide which parts of its learned belief remain useful. The
switch rounds $\{s_i\}$ are also staggered across customers. No single time step
marks the shift, forcing the focal merchant to diagnose each customer
individually from interleaved post-shift evidence. Section~\ref{sec:setup-shock}
specifies the transformation $\mathcal{T}_i$ used in our experiments.

\subsection{Diagnostics: Surplus and Oracle Regret}
\label{sec:model-surplus}

Let $f\in\mathcal{M}$ denote the focal merchant. We first compute zero-margin
surplus and the corresponding structural surplus gap:
\begin{equation}
  \small
\begin{aligned}
  S_{i,m,t}(\mathbf{x}) &= V_{i,t}(\mathbf{x}) - c_m(\mathbf{x}),\quad
  S^*_{i,m,t}=\max_{\mathbf{x}}S_{i,m,t}(\mathbf{x}), \\
  \Delta_{i,m,t} &= S^*_{i,m,t}-\max_{j\ne m}S^*_{i,j,t}.
\end{aligned}
\end{equation}
Positive $\Delta_{i,m,t}$ means customer $i$ is structurally favorable to
merchant $m$ under cost-price competition. Because these quantities depend on
hidden values and costs, merchants cannot observe them; we use them only for
evaluation.

For hindsight-oracle regret, define the competitor utility to beat, $q_{i,t}$,
and the corresponding oracle profit, $\pi^{\star}_{i,t}$:
\vspace{-0.5em}
\begin{equation}
\begin{aligned}
  q_{i,t} &= \max_{j\ne f}\bigl[V_{i,t}(\mathbf{x}_{j,i,t}) - p_{j,i,t}\bigr], \\
  \pi^{\star}_{i,t} &= \max_{\mathbf{x}}
  \left[V_{i,t}(\mathbf{x}) - q_{i,t} - c_f(\mathbf{x})\right]_+,
\end{aligned}
\label{eq:oracle-profit}
\end{equation}
where $[z]_+=\max(z,0)$. Comparing $\pi^{\star}_{i,t}$ to realized profit
separates two errors: losing a profitably winnable customer-round, and winning
but underpricing. Full metric definitions are in Section~\ref{sec:setup-metrics}.
\section{Experimental Setup}
\label{sec:setup}
We instantiate the framework in Section~\ref{sec:model} as a repeated
multi-attribute auction with one LLM-controlled merchant and three adaptive rule-based opponents (bots). The design keeps the game small enough to run
across many frontier models and seeds, while still stress-testing three capabilities: customers preference learning, margin management, and adaptation after unannounced shifts.

\noindent\textbf{Objective.}
The agent's goal is to maximize total profit throughout the game.

\subsection{Environment Instantiation}
\label{sec:setup-instantiation}

A detailed version of our market, customer-level auction, and learning loop
appears in Figure~\ref{fig:overview} (Appendix~\ref{app:overview}). We
use three abstract attributes, denoted $A$, $B$, and $C$, with five levels each, giving $5^3=125$ configurations. The
focal merchant is a generalist: it can combine all three attributes in one
offer. The opponents are specialists, each structurally advantaged on one
attribute. Attribute names are intentionally abstract to avoid giving LLMs
domain priors from labels such as storage capacity, screen size, or material
grade. Likewise for customers and merchants so the agent cannot anchor on
persona stereotypes or branded retailers.

With attribute base costs equalized,
\begin{equation}
  C_A=C_B=C_C=[0,1,3,6,12],
\end{equation}
no attribute is inherently cheaper to provide; any structural advantage
comes from merchant specialization. Concretely (Eq.~\ref{eq:cost-general}),
$C_{m,k}[\ell]=\mu^m_k C_k[\ell]$ with all merchants sharing base cost
$c_0=70$. The focal generalist has multiplier $0.75$ on every attribute, and
each specialist has $0.4$ on its specialty and $1.0$ on the others
(Table~\ref{tab:merchants}).

\begin{table}[b]
\centering
\resizebox{0.85\columnwidth}{!}{%
\begin{tabular}{lcccc}
\toprule
\textbf{Merchant} & $\mu_A$ & $\mu_B$ & $\mu_C$ & \textbf{Role} \\
\midrule
Focal LLM & 0.75 & 0.75 & 0.75 & Generalist \\
$A$-specialist bot & 0.4 & 1.0 & 1.0 & Specialist \\
$B$-specialist bot & 1.0 & 0.4 & 1.0 & Specialist \\
$C$-specialist bot & 1.0 & 1.0 & 0.4 & Specialist \\
\bottomrule
\end{tabular}
}
\caption{Cost multipliers in our instantiation.}
\label{tab:merchants}
\end{table}

\subsection{Customers and Preference Shocks}
\label{sec:setup-shock}

On the demand side, we use 24 customers, each with hidden value curves over
the five levels of $A$, $B$, and $C$. The mix spans three competitive
regimes: customers where one attribute dominates (4 \emph{singles},
structurally favoring specialists), customers split between two (12
\emph{duals}), and customers rewarding full bundles (8 \emph{triples}).
Appendix~\ref{app:customers} gives the value-curve shapes, the construction
procedure, and the full customer list.

Twelve customers receive the preference shock: four $A\leftrightarrow B$
swaps, four $A\leftrightarrow C$, and four $B\leftrightarrow C$. The
remaining twelve serve as controls. Each shocked customer's shock round is
sampled uniformly from $\{31,\ldots,40\}$ and fixed by the experiment seed.
A swap exchanges two value curves and leaves the third unchanged, preserving demand while changing the preferred configuration.

\subsection{LLM Agents and Tools}

The LLM agent plays a single focal merchant against the 24 customers across
repeated rounds, generating a bid for each customer in each round. After
every customer-level auction, it writes a belief update for
that customer's preferences, conditioned on the win/lose outcome and the revealed winning offer. After all 24 auctions in a round resolve, it writes one global strategy update. At bid stage, the agent may call tools that expose its own state, cost structure (Sec.~\ref{sec:setup-instantiation}), and past beliefs, but no hidden customer values or competitor state.

\paragraph{Per-round outputs.} The agent produces (i) 24 customer bids of the form $(a,b,c)\,@\,\$p$, e.g.\ $(3,2,1)\,@\,\$84$; (ii) 24 customer-level natural-language belief updates; and (iii) one global strategy update at the end of the round.

\subsection{Opponent Dynamics}
\label{sec:setup-bots}

Each specialist bot always offers its specialty configuration: $(4,0,0)$
for the $A$-specialist bot, $(0,4,0)$ for $B$-specialist, and $(0,0,4)$ for $C$-specialist. Bots adapt only price. For each customer, a bot opens at a $\$3$
margin over cost, raises its customer-specific margin after a win by a step
$\sim\mathcal{U}(0.5,1.5)$, and lowers it after a loss by a step
$\sim\mathcal{U}(0.25,0.75)$, floored at $\$1$ over cost. The asymmetric raise/lower distributions ensure bots climb margin faster than they cede it, which prevents trivial exploitation by an agent that simply underbids once.
Bots randomness is seeded by the experiment seed. Appendix~\ref{app:bots}
gives the full rule.

This opponent design is simple but nontrivial. Prior LLM auction benchmarks
often use rule-based reference agents, such as fixed-increment bidding in
AucArena~\cite{chen2023aucarena}, or study LLMs as auction participants without
adaptive non-LLM pricing opponents~\cite{shah2025syntheticlabs,yin2025infobid}.
Our specialists are stronger than static-price opponents because they adapt
margins separately for each customer, yet they remain interpretable. Product
choice is fixed by structural cost advantage, and only price changes over time.
This creates a controlled test of whether the LLM generalist can identify where
bundled offers are valuable, avoid unprofitable overcompetition, and extract margin when it has an advantage.

\subsection{Execution Protocol}
\label{sec:setup-protocol}

A run proceeds in three temporal segments. In rounds 1--30 all 24 customers
hold their initial preferences. We chose this 30-round pre-shock window
empirically across our model sweep, the slowest models reach saturation
within roughly this many rounds. In rounds 31--40 the 12 shocked customers undergo their
paired-attribute swap at the customer-specific round drawn in
\S\ref{sec:setup-shock}, while controls are unaffected. Each customer then runs
for 40 further rounds past its own shock round (or past round~40 if
unshocked), giving the agent time to detect and adapt.

Within each round the four merchants submit bids simultaneously, with no
merchant seeing another's offer. The customer selects the offer that maximizes its utility, and the auction reveals the winning offer and merchant to the agent. 
To account for LLM stochasticity, every model is evaluated on ten random seeds, which also fix bot randomness and
the shock schedule. Per-stage prompts and the session architecture are
detailed in Appendices~\ref{app:prompts} and~\ref{app:tokens}.

\subsection{Bandit Baselines}
\label{sec:setup-bandits}

We compare against Thompson Sampling~\cite{thompson1933} and
EXP4~\cite{auer2002exp4} baselines on the same
configuration-and-margin action space, each instantiated three ways: a separate bandit per customer, 
one shared per customer type-cluster, or a single global bandit. The
bandits face the same customer population, opponent dynamics, and shock
schedules as the LLMs, and consume the same binary win/loss signal. Update
rules, the factored action space, and exploration parameters are detailed in
Appendix~\ref{app:baselines}.

\subsection{Metrics and Statistical Methodology}
\label{sec:setup-metrics}

We report win rate and total profit over rounds the focal won, and margin per win, and margin per win averaged across seeds as standard coverage and pricing summaries. Shock
recovery is the agent's win rate in a late post-shock window (last 10 rounds) minus its peak pre-shock win-rate. We also define the following met:ics:
\begin{itemize}[leftmargin=*,itemsep=1pt]
  \item \textbf{Oracle efficiency}: realized profit divided by
    hindsight-oracle profit (Eq.~\ref{eq:oracle-profit}), representing the maximum profit the agent could have earned if it knew each customer's preferences and charged the optimal price.
    \item \textbf{Regret decomposition.} The gap to the oracle
    
    splits into
    \emph{loss regret} (rounds the agent lost but the oracle would have won)
    and \emph{underprice regret} (rounds the agent won but priced below the
    oracle optimal).
  \item \textbf{Robustness}: The oracle ceiling above is endogenous: a stronger agent forces the bots to tighten margins, which compresses its own ceiling. To remove this dependence, we recompute the oracle replacing each round's actual
    competitor utility with the across-model median.
\end{itemize}

\section{Results}
\label{sec:results}

We evaluate eleven frontier LLMs from four providers as focal merchants. All
models face the same 24 customers, three specialist bot opponents, and a
staggered preference shock. We report profit, win rate, and margin per win
averaged over ten seeds; per-seed results are in
Appendix~\ref{app:per-seed}.

\subsection{Main leaderboard}
\label{sec:results-leaderboard}

Table~\ref{tab:leaderboard} presents the full results. 
Pairwise permutation tests (10{,}000 shuffles) identify four statistically distinct tiers; models within a tier are not separable at significance threshold  $\alpha = 0.05$.

\begin{table}[t]
\centering
\footnotesize
\setlength{\tabcolsep}{2pt}
\renewcommand{\arraystretch}{0.94}
\begin{tabular}{@{}rllrrr@{}}
	\toprule
	\textbf{\#} & \textbf{Model} & \textbf{Think/Effort} & \textbf{Profit (\$)} & \textbf{Win\%} & \textbf{\$/Win} \\
\midrule
\multicolumn{6}{@{}l}{\cellcolor{gray!8}\textit{Tier 1}} \\
1  & Opus 4.6 & adapt/xhigh & \textbf{2{,}976}\,$\pm$354 & 67.8 & \textbf{2.39} \\
2  & Opus 4.6 & off/high & 2{,}769\,$\pm$425 & 67.8 & 2.22 \\
3  & GPT-5.5$^\ddagger$ & --/xhigh & 2{,}693\,$\pm$715 & 77.4 & 1.91 \\
4  & Gemini Pro & dyn/high & 2{,}614\,$\pm$378 & \textbf{77.9} & 1.83 \\
5  & Opus 4.6 & adapt/med & 2{,}570\,$\pm$343 & 66.8 & 2.10 \\
\addlinespace[1pt]
\multicolumn{6}{@{}l}{\cellcolor{gray!8}\textit{Tier 2}} \\
6  & GPT-5.4 & --/high & 1{,}936\,$\pm$203 & 76.0 & 1.38 \\
7  & GPT-5.5 & --/high & 1{,}705\,$\pm$563 & 77.4 & 1.20 \\
8  & Opus 4.5 & off/high & 1{,}677\,$\pm$388 & 66.6 & 1.37 \\
9  & Gemini Flash & dyn/high & 1{,}660\,$\pm$367 & 65.2 & 1.39 \\
10 & Grok 4.2 & --/-- & 1{,}554\,$\pm$612 & 57.7 & 1.44 \\
11 & GPT-5.3 & --/-- & 1{,}536\,$\pm$363 & 66.4 & 1.26 \\
12 & Opus 4.7 & adapt/xhigh & 1{,}492\,$\pm$298 & 64.5 & 1.26 \\
13 & Sonnet 4.6 & adapt/med & 1{,}407\,$\pm$170 & 55.9 & 1.37 \\
14 & Sonnet 4.6 & off/high & 1{,}306\,$\pm$343 & 56.3 & 1.26 \\
15 & Opus 4.7 & adapt/med & 1{,}222\,$\pm$348 & 55.6 & 1.20 \\
16 & Grok 4.1 & --/-- & 1{,}197\,$\pm$444 & 57.8 & 1.13 \\
\addlinespace[1pt]
\multicolumn{6}{@{}l}{\cellcolor{gray!8}\textit{Tier 3}} \\
17 & Opus 4.7 & off/high & 1{,}001\,$\pm$455 & 55.1 & 0.99 \\
18 & GPT-5.5 & --/med & 941\,$\pm$308 & 61.8 & 0.83 \\
19 & GPT-5.4 & --/med & 821\,$\pm$302 & 61.7 & 0.72 \\
\addlinespace[1pt]
\multicolumn{6}{@{}l}{\cellcolor{gray!8}\textit{Tier 4}} \\
20 & Grok 4.1 NR & --/-- & 544\,$\pm$323 & 45.1 & 0.66 \\
21 & GPT-5.4 & --/none & 266\,$\pm$196 & 36.2 & 0.40 \\
\midrule
\multicolumn{6}{@{}l}{\cellcolor{gray!8}\textit{Bandit baselines}} \\
   & TS & --/-- & 1{,}426\,$\pm$101 & 66.4 & 1.17 \\
   & EXP4 & --/-- & 1{,}023\,$\pm$26 & 35.5 & 1.57 \\
\bottomrule
\end{tabular}
\caption{Full-game leaderboard across 11 frontier LLMs and their variants,
averaged over 10 seeds. Profit is cumulative over ${\sim}79$ rounds; \$/Win is
mean realized margin per winning bid. \textbf{Think/Effort}: the first item is
thinking mode and the second is provider-specific effort level. adapt =
adaptive, off = thinking disabled, dyn = Google's dynamic thinking mode, med =
medium, and -- = N/A. $^\ddagger$3 seeds
rather than 10.}
\label{tab:leaderboard}
\end{table}

\begin{figure}[t]
\centering
\includegraphics[width=0.95\linewidth]{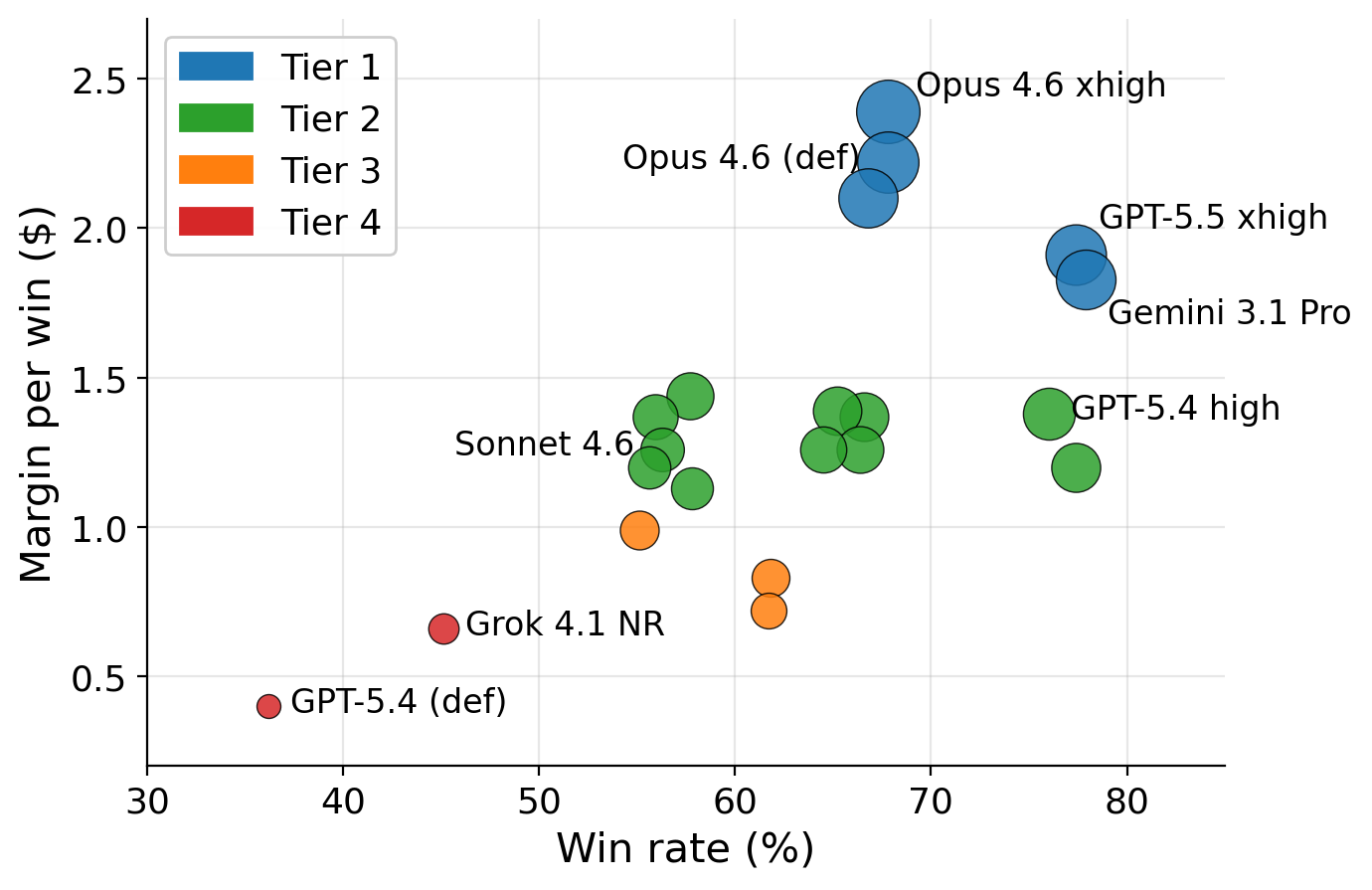}
\caption{Profit is driven by margin, not coverage. Each point is a
variant; bubble size is proportional to profit, and color
indicates leaderboard tier. Unlabeled points correspond to other
variants listed in Table~\ref{tab:leaderboard}.}
\label{fig:winrate-margin}
\end{figure}

\paragraph{Bazaar differentiates but is not trivially solved.}
Total profit spans an $11{\times}$ range, from \$2{,}976 for Opus~4.6
at xhigh effort to \$266 for GPT-5.4 at none. The gap is not just a
thinking-budget artifact: even at provider defaults the spread remains
${\sim}3{\times}$ (\$2{,}769 for Opus~4.6 vs.\ \$941 for GPT-5.5,
excluding GPT-5.4's outlier no-effort default). \sysname{} thus
imposes enough competitive pressure to stratify strong models rather than saturating at the top of the leaderboard.

\paragraph{Margin discipline drives profit more than win rate.}
Across models, total profit correlates with margin per win at $r=0.99$
but with win rate at only $r=0.88$ (Figure~\ref{fig:winrate-margin}).
Gemini~3.1~Pro wins 10\% more often than
Opus~4.6 (77.9\% vs.\ 67.8\%) yet earns less total profit, making \$1.83 per win versus \$2.22 for Opus~4.6. Winning more auctions is
not enough if the model underprices. The benchmark rewards
\emph{knowing how much to charge}, not merely \emph{what to offer}.

\paragraph{Thinking effort strongly affects profit.}
Holding the base model fixed, GPT-5.4 earns \$266 at default (Tier~4)
but \$1{,}936 at high effort (Tier~2), a $7.3{\times}$ increase.
GPT-5.5 rises from \$941 to \$1{,}705, and Opus~4.6 improves from
\$2{,}769 to \$2{,}976. Within-family reasoning-budget gains exceed several between-family gaps, making inference-time compute a first-order determinant of agent performance. However, the dollar API spend required to run these higher effort variants varies sharply across models (Appendix~\ref{app:tokens}). Notably, Gemini~3.1~Pro is the most cost-efficient top performer, \#2 by profit but $4.8\times$ cheaper than Opus~4.6 (\$186 vs.\ \$890 API cost per seed).

\paragraph{Bandits set a non-trivial floor.}
Per-customer Thompson Sampling earns \$1{,}426, above six of eleven
base LLMs. The advantage of top LLMs appears in margin (Opus~4.6:
\$2.22 per win vs.\ \$1.17 for Thompson Sampling) and, as we show in
\S\ref{sec:results-shock}, in the \emph{speed} of post-shock adaptation.

\subsection{Shock adaptation by type}
\label{sec:results-shock}

During rounds 31--40, half of the customers undergo paired-attribute
preference swaps. Although the cost structure is symmetric across attributes, recovery still varies by swap type (Table~\ref{tab:shock-recovery}), reflecting how each shift interacts with the model's learned customer beliefs and prior pricing history. The win-rate trajectory on shocked customers (Figure~\ref{fig:shock-traj}) shows full recovery within 10--15 rounds for the best LLMs and persistent degradation of 20\% or more for the worst.

\begin{table}[t]
\centering
\small
\resizebox{\columnwidth}{!}{%
\begin{tabular}{lcccc}
\toprule
Model & A$\leftrightarrow$B & A$\leftrightarrow$C & B$\leftrightarrow$C & Avg \\
\midrule
Gemini 3.1 Pro       &  $\phantom{+}0.0$ & $\mathbf{+11.2}$ &  $-1.2$ & $\mathbf{+3.3}$ \\
Grok 4.1 Reasoning   &  $+4.2$ & $+8.3$ &  $-8.3$ & $+1.4$ \\
Gemini 3 Flash       &  $\phantom{+}0.0$ & $-4.2$ &  $-5.0$ & $-3.1$ \\
Claude Opus 4.6      & $-14.2$ &$-23.3$ & $\mathbf{+1.7}$ & $-11.9$ \\
Claude Sonnet 4.6    & $-11.7$ &$-11.7$ & $-16.7$ & $-13.4$ \\
Claude Opus 4.5      & $-21.7$ &$-12.5$ & $-23.3$ & $-19.2$ \\
GPT-5.3              & $-10.8$ &$-22.5$ & $-24.2$ & $-19.2$ \\
\bottomrule
\end{tabular}
}
\caption{Shock recovery by type: pp change from pre-shock peak (R26--30) to the final post-shock window (S$+30$--$39$). Positive means recovery; negative
means degradation.}
\label{tab:shock-recovery}
\end{table}

\paragraph{Strong-learner / weak-adapter trade-off.} We measure pre-shock learning speed as the rise in win rate from round~1 to the R26--30 peak window (full curves in Appendix~\ref{app:learning-curves}). The models that climb fastest during this window — GPT-5.3 (+46~pp) and Claude Opus~4.5 (+40~pp) — are among the worst adapters, losing $\sim$19 points on average across shock types. Gemini~3.1~Pro shows the opposite pattern: it plateaus earlier pre-shock but exceeds its pre-shock performance on $A\leftrightarrow C$ shocks by +11.2 points. This suggests that strong pre-shock beliefs can be hard to revise, while Gemini's concise, hypothesis-oriented strategies (\S\ref{sec:results-reasoning}) leave more room for re-inference.

\paragraph{Bandits recover slowly.}
Per-customer Thompson Sampling recovers to its pre-shock 78\% win rate only by round $s{+}30$, while the best LLMs recover within 5--10 rounds. This
$3{-}5\times$ speed advantage translates into higher cumulative profit during
the recovery window (Appendix~\ref{app:per-seed}).

\subsection{Reasoning behavior}
\label{sec:results-reasoning}

The free-form strategy text emitted at each round lets us look beyond
scores at \emph{how} models reason. Three behavioral archetypes emerge
across models (qualitative excerpts in
Appendix~\ref{app:strategy-style}). \emph{Structural reasoners} write
concise, hypothesis-shaped strategies: Gemini~3.1~Pro produces a
median of 325 characters by round~30, against 1{,}400+ for most other
models. \emph{Lock-in agents} produce verbose, commitment-encoded prose
with explicit self-instructions not to deviate, as when Opus~4.6 records
a per-customer configuration/price ledger headed
``ABSOLUTELY IDENTICAL, DO NOT CHANGE.'' \emph{Pruners} aggressively
drop customers altogether: Grok~4.1 Reasoning bids on only 8 of 24
customers by round~40.

These archetypes do not simply reproduce the leaderboard. Opus~4.6 is closer in z-scored behavioral space to Gemini~3.1~Pro, its top rival, than to Opus~4.7, its same-family successor. Pricing skill is therefore not just a matter of rank: it reflects both how a
model reasons and how well that reasoning translates into bids.

Zero-bid intent classification (Appendix~\ref{app:giveup}) sharpens
the picture. A \emph{zero bid} is $(0,0,0)$ priced at cost. The same
surface action means different things
across archetypes: Sonnet~4.6's zero bids are mostly explicit forfeits
(59\%), Opus~4.7's are strategic floor-setting (32\%), and Grok's are
exploratory probes (40--60\%).

Traces show a gap between detection and revision. Strategy and belief
text (excerpts in Appendix~\ref{app:reasoning-traces}) show that models
register losses within a few rounds of a shock. Models that fail to
recover then commit to a single revised hypothesis and iterate within it.
Revision, not detection, is the bottleneck in post-shock adaptation.

\subsection{Regret analysis}
\label{sec:results-regret}

Raw profit does not reveal how close a model came to optimal play
given the competitive responses it provoked. Following standard online-pricing
work~\cite{kleinberg2003value,besbes2009dynamic,denboer2015dynamic},
we measure \emph{regret efficiency} (defined in Sec.~\ref{sec:setup-metrics}), denoted by $\eta$. Because bots adapt to the focal model's past
play, the ceiling is endogenous; this reflects the
\emph{policy-regret} insight~\cite{arora2012policy} that a stronger
agent compresses its own ceiling. Total regret decomposes additively
into \emph{loss regret} (oracle profits forfeited in rounds the focal
lost) and \emph{underprice regret} (margin left on the table in rounds
the focal won), a split analogous to the missed-sales / margin-erosion
decomposition in revenue management~\cite{talluri2004revenue}.

\paragraph{Efficiency reorders the leaderboard.}
Table~\ref{tab:regret} reports $\eta$ alongside the regret
decomposition. Gemini~3.1~Pro achieves the highest base-model
efficiency (0.310), overtaking Opus~4.6 (0.295) despite earning less
total profit, because it wastes less of a smaller oracle ceiling.
Among thinking-augmented variants, Opus~4.6 with extended thinking
(xhigh) has the highest efficiency at 0.321.

\begin{table}[t]
\centering
\small
\setlength{\tabcolsep}{4pt}
\resizebox{\columnwidth}{!}{%
\begin{tabular}{@{}rlrrr@{}}
\toprule
\# & \textbf{Model (thinking, effort)} & $\eta$ & \textbf{Under} & \textbf{Loss} \\
\midrule
1  & Opus 4.6 (adaptive, xhigh)     & \textbf{0.321} & \$3{,}628 & \$2{,}678 \\
2  & Gemini 3.1 Pro (dynamic, high) & 0.310 & \$4{,}264 & \$1{,}565 \\
4  & Opus 4.6 (off, high)           & 0.295 & \$3{,}769 & \$2{,}861 \\
6  & GPT-5.4 (high)                 & 0.225 & \$5{,}058 & \$1{,}570 \\
8  & Opus 4.5 (off, high)           & 0.181 & \$4{,}332 & \$3{,}297 \\
16 & Opus 4.7 (off, high)           & 0.100 & \$3{,}945 & \$5{,}197 \\
19 & GPT-5.4 (none)                 & 0.025 & \$3{,}054 & \$7{,}703 \\
\bottomrule
\end{tabular}
}
\caption{Regret efficiency $\eta$ = realized/oracle
profit; Under = margin left on won rounds; Loss = oracle profit forfeited on
lost rounds (total regret = Under + Loss). Full
leaderboard with standard errors in Appendix~\ref{app:regret-full}.}
\label{tab:regret}
\end{table}

\begin{figure}[t]
\centering
\includegraphics[width=\linewidth]{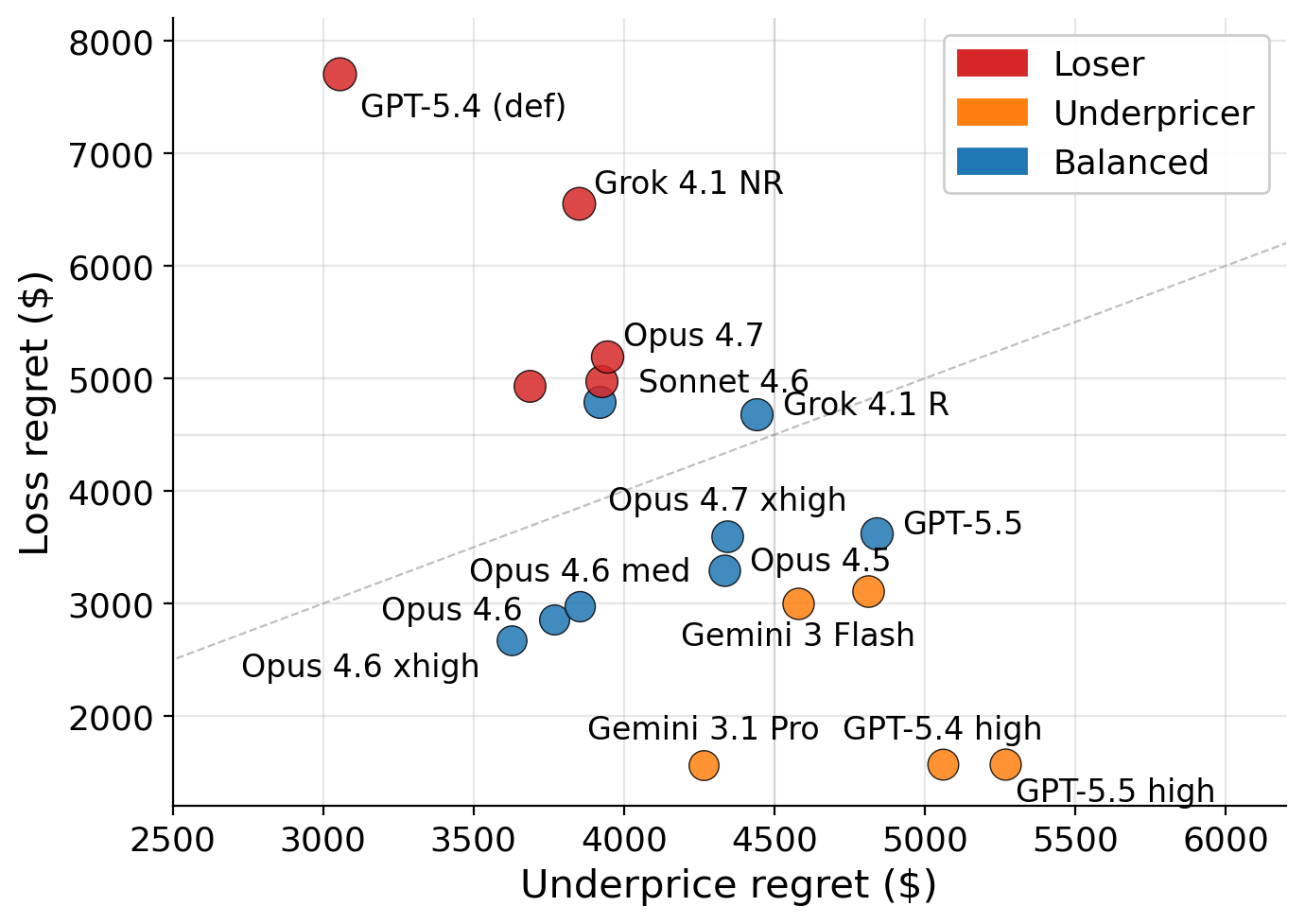}
\caption{The regret plane.
Top-left
\emph{Losers} forfeit auctions the oracle would win; bottom-right
\emph{Underpricers} win but leave surplus on the table.}
\label{fig:archetype-scatter}
\end{figure}

\paragraph{Models fail in qualitatively different ways.}
The loss/underprice decomposition (Figure~\ref{fig:archetype-scatter})
reveals two distinct failure archetypes. \emph{Losers} (GPT-5.4 base,
Grok~NR, Opus~4.7, Sonnet~4.6) forfeit the majority of their regret
by losing auctions the oracle would have won: they bid too low or on
the wrong configuration, and competitors take the customer entirely.
\emph{Underpricers} (Gemini~3.1~Pro, GPT-5.4 high, GPT-5.5 high) win
most auctions they should but consistently leave surplus on the
table; they could have charged more without losing the customer. The
top models (Opus~4.6, Opus~4.5) are balanced, splitting regret roughly
evenly between the two sources. Per-variant loss/underprice splits
for all 19 entries are in Appendix~\ref{app:regret-full}.

\paragraph{Thinking budget reshapes the mistake profile.}
Extended inference compute does not reduce all regret uniformly; it
targets loss regret specifically. GPT-5.4 at default sources 72\% of
its regret from losses; at high effort that fraction collapses to
24\% (a $4.9\times$ reduction), while underprice regret
\emph{increases} from \$3{,}054 to \$5{,}058. The model transforms
from a Loser into an Underpricer. GPT-5.5 follows the same pattern
(43\%~$\rightarrow$~23\% loss fraction; see
Appendix~\ref{app:regret-full}); Opus~4.6, already balanced at
default, gains efficiency without shifting archetype. Per-variant
regret breakdowns across the thinking-effort ladder are in
Appendix~\ref{app:effort-deltas}.

\paragraph{Newer is not always better.}
At provider defaults, the newest base model in three of four
families regresses against an earlier version (full numbers in
Appendix~\ref{app:regret-full}). Opus~4.7 (0.100) trails both
Opus~4.6 (0.295) and Opus~4.5 (0.181). Among OpenAI base models GPT-5.3 (0.163)
outperforms both GPT-5.4 (0.025) and GPT-5.5 (0.100). Only xAI bucks
the trend: Grok~4.2 reasoning (0.152) improves on Grok~4.1 reasoning
(0.116). Extended thinking narrows or reverses these gaps:
GPT-5.5~xhigh (0.307) clears GPT-5.3, and Opus~4.7~xhigh (0.158)
closes most of the deficit to 4.5. On this benchmark, version recency
is a poor predictor of out-of-the-box competence.

\paragraph{Robustness.}
The ranking is robust to a fixed-competitor oracle that holds bot
behavior constant; the only swap in the top-7 is
Gemini~3.1~Pro and Opus~4.6 exchanging positions 1--2 (fixed-oracle:
Opus~4.6 = 0.293, Gemini~3.1~Pro = 0.276).

\subsection{Efficiency Analysis}
\label{sec:analysis}

\paragraph{Reasoning text.}
Our architecture maintains separate conversation sessions per customer
per agent (Appendix~\ref{app:tokens}), so we require models to output
explicit reasoning each round. Providers do not consistently and reliably expose internal
thinking tokens in a portable form. Anthropic offers only a summary,
Google's thinking is inaccessible through the OpenAI-compatible API we
use, and OpenAI and xAI surface token counts only. We therefore require
\texttt{reasoning}, \texttt{target\_belief}, and
\texttt{new\_strategy} fields in structured output. These fields are
reused in subsequent rounds (beliefs retrieved via tool call, strategy
accessible from any session, reasoning piped into the round-strategy
update), and we treat their quality and length as a proxy for how
effectively the model maintains state.

\paragraph{Strategy-note verbosity.} Mean per-round strategy-text
length varies from 236~chars
(Grok~4.1~Reasoning) to 1{,}760~chars (Sonnet~4.6).
Gemini~3.1~Pro averages 534~chars, roughly a third of Opus~4.6's
1{,}734~chars (adaptive, xhigh), yet is one of the top performing models.
Most models' notes shorten as they converge: Opus~4.6 (adaptive, xhigh)
begins near 2{,}200~chars and settles around 1{,}300 by the final ten rounds,
a 1.7$\times$ decline that holds across all its thinking variants.
Gemini~3.1~Pro stays essentially flat
throughout (553$\to$473 chars, a 15\% decline over 80 rounds).
Opus~4.7 (adaptive, xhigh) is an outlier with notes declining from
1{,}087~chars in the first ten rounds to 369 by the final
ten-round window, a $3\times$ reduction. At lower effort levels the collapse is more severe (up to
$7\times$ at default high).

Token spend ($2.3\times$ across models) does not track the leaderboard,
and estimated per-seed API cost varies $14\times$ (driven primarily by
per-token pricing differences across providers): Gemini~3.1~Pro is the
most cost-effective by a wide margin, while the most expensive
model ranks mid-table. Full breakdowns are in
Appendix~\ref{app:tokens}.

\section*{Limitations}
Our customers are \emph{synthetic}: their preference vectors are
drawn from hand-constructed archetype clusters rather than real
consumer data. This is deliberate (it gives us oracle-level ground
truth for defining surplus, competitive advantage, and shock
magnitude), but the specific numeric gaps we report should be
read as relative signals rather than absolute deployment forecasts.
Second, each run pairs one LLM against three identical bots, so
the bandit and LLM merchants face different competitive
environments from what multi-LLM competition would present; we
treat the bot environment as a controlled probe of pricing skill,
not a simulation of live markets. Third, our analysis of reasoning
style is qualitative; a systematic study of how prompt structure
and scratchpad organization affect pricing and adaptation is left
to future work. Finally, the preference shock is a single, large,
intra-experiment disturbance; smooth drift and repeated shocks
remain open.

\section*{Ethics Statement}
\label{sec:ethics}

\paragraph{Per-customer pricing and value-based differentiation.}
The pricing behavior \sysname{} measures, a merchant choosing a
different product configuration and price for each customer, resembles personalized price discrimination, a practice that is ethically and legally fraught when it is linked to demographic or socioeconomic attributes such as race, gender, location, income, or device fingerprint. The mechanism we study is categorically different. The agent's input is the \emph{product preference} of each customer, expressed as the win/loss outcome of past competitive
auctions: which attribute bundle the customer chose, and at what
price relative to competing offers. Our setup exposes only an opaque target identifier and the history of attribute choices for that identifier. The decision the agent is making is therefore which \emph{product variant} (e.g., higher warranty and slower shipping vs.\ basic warranty and expedited shipping) best fits this customer's previously expressed needs, and what margin to attach to it.

This is the same ethical structure as enterprise B2B contract pricing,
where a vendor learns over repeated negotiations that one customer
values service-level guarantees and another values onboarding support,
and assembles a package and price accordingly; or as tiered SaaS
plans, where customers self-select into bundles that match their usage.
In both cases customers receive different prices because they receive
\emph{different products}, not because of who they are, and they
retain the option of competing offers.

\paragraph{Risks if deployed without guardrails.}
Even on attribute-only inputs, customer clusters can correlate
with protected groups in real markets, so deployments should monitor
disparate pricing outcomes across protected attributes. Our benchmark deliberately omits identity features; for example customers, merchants, and attributes are given abstract numerical IDs so that the failure modes we report are attributable to pricing skill rather than to identity-based
discrimination.

\paragraph{Data and participants.}
The customer population in our experiments is fully synthetic:
preference vectors are sampled from hand-constructed archetype
clusters (\S\ref{sec:setup}) and contain no information traceable to
real individuals. No human-subjects data was collected and no
deployed pricing system was instrumented in the production of these
results.

\paragraph{Compute and reproducibility.}
LLM inference accounts for essentially all of the compute footprint of
this work. We report per-cell API token usage and list-price cost
estimates (Appendix~\ref{app:tokens}) so that practitioners can weigh
the environmental and monetary cost of replicating or extending the
benchmark. Reproducible artifacts (prompts, tool schemas, per-seed
run logs, and behavioral traces) are released alongside the paper to
allow independent verification without re-running the full sweep.

\bibliography{refs}

\appendix
\section{Related Work}
\label{app:related}

Our work builds on three threads: LLM agents in strategic and
commerce settings, prior LLM-for-commerce benchmarks on the buyer
side, and the algorithmic-pricing literature with its regret-based
evaluation tradition.

\paragraph{LLM agents in auctions, negotiation, and multi-agent commerce.}
A growing body of work places LLMs into structured strategic
interactions, and the auction setting has been a natural starting
point. AucArena~\cite{chen2023aucarena} runs LLM bidders in English
auctions and reports large capability gaps between models, though its
only opponent is a fixed-rule baseline. Shah et
al.~\cite{shah2025syntheticlabs} take a different angle and ask
whether LLMs reproduce known behavioral regularities such as risk
aversion and sniping when used as synthetic subjects in sealed-bid
auctions, while InfoBid~\cite{yin2025infobid} probes how information
disclosure shifts what LLM bidders do. A parallel thread looks at
dialogue rather than bidding.
NegotiationArena~\cite{pmlr-v235-bianchi24a} and
GLEE~\cite{shapira2024glee} evaluate multi-turn buy and sell
negotiations across LLMs, and GLEE in particular contributes a large
LLM-to-LLM and LLM-to-human dataset that spans bargaining,
negotiation, and persuasion, building on earlier work like
\cite{he2018decoupling}. Measuring
Bargaining~\cite{measuringbargain2024} grounds the same question in
real Amazon price-negotiation data, and the work in
\cite{evalmultiturn2025,gametheory24} examines intent recognition in
buyer and seller dialogues alongside structured workflows for
strategic reasoning. Broader multi-agent benchmarks like
Magentic~\cite{magnetic25}, MAGPIE~\cite{juneja2025magpie}, and
NegotiationToM~\cite{negotiationtom24} push further out, covering full
search-to-payment commerce lifecycles, private-information leakage,
and theory-of-mind. Two patterns recur across all of this. The
strategic medium is dialogue, so capability shows up through messages
rather than prices, and the evaluation pool is essentially static,
with fixed scripts, fixed valuations, and no preference shifts.
\sysname{} departs from both by being bid-based and explicitly
competitive. It pairs LLMs against adaptive specialist opponents and
classical bandit baselines, and stresses adaptation by injecting a
mid-experiment preference shock.

\paragraph{LLMs in e-commerce (buyer- and recommendation-side).}
A separate thread evaluates LLMs on consumer-facing e-commerce tasks,
and most of it sits firmly on the buyer side.
WebShop~\cite{yao2022webshop} casts product search and purchase as a
sequential decision problem, while eCeLLM~\cite{peng2024ecellm}
instruction-tunes LLMs to handle the underlying primitives of product
attribute extraction, query understanding, and sequential
recommendation. ChineseEcomQA~\cite{chen2025chineseecomqa} steps back
and offers a large concept benchmark aimed at grounding e-commerce
knowledge in the first place. Closer to deployed assistants, Shopping
Companion~\cite{yu2026shoppingcompanion} introduces a memory-augmented
buyer-side agent, and Automated but Risky
Game~\cite{zhu2025automatedrisky} documents what can go wrong when
LLMs transact with each other, including overspending and anomalous
purchases. The common thread is that these works target the
user-experience and product-understanding side of commerce rather than
the seller-side pricing question we study. We note for completeness
that multi-attribute or scoring auctions, where bids combine price
with non-price attributes~\cite{asker2008scoring}, are a closer formal
cousin to our setting but address a different selection problem.

\paragraph{Algorithmic pricing and regret-based evaluation.}
The standard evaluation framework for online pricing is cumulative
\emph{regret} measured against a hindsight-optimal oracle. The framing
was established by Kleinberg and
Leighton~\cite{kleinberg2003value} for posted-price auctions, extended
to continuous pricing under unknown demand by Besbes and
Zeevi~\cite{besbes2009dynamic}, and laid out comprehensively in the
survey of Den Boer~\cite{denboer2015dynamic}. We borrow Thompson
Sampling~\cite{thompson1933} and EXP4~\cite{auer2002exp4} from this
tradition and run them on the same action space as our LLM agents.
One subtlety is worth flagging. Standard external regret compares
against a fixed-action oracle, but when the environment adapts to the
agent's past play the right notion becomes \emph{policy
regret}~\cite{arora2012policy}, so our hindsight oracle is itself
endogenous and is computed accordingly in
\S\ref{sec:results-regret}. Closer to our multi-agent setting,
Balseiro and Gur~\cite{balseiro2019learning} apply regret minimization
to budget-constrained repeated auctions. On the LLM side, Fish et
al.~\cite{fish2024algorithmiccollusion} show that prompt-only LLM
merchants can drift into supra-competitive equilibria, echoing the
classical Q-learning collusion result of Calvano et
al.~\cite{calvano2020aipricingcollusion}. We do not target collusion
in this paper, but the repeated and adaptive structure of \sysname{}
is exactly the kind of setting in which it could surface, and we view
it as a natural follow-up. Our loss-versus-underprice decomposition
adapts the missed-sales and margin-erosion distinction from revenue
management~\cite{talluri2004revenue}.

Taken together, our contribution is to put modern LLM agents inside
this regret-based tradition, equipped with bandit baselines, an
explicit preference shock, and behavioral diagnostics that surface
failure modes such as margin-at-cost bidding, give-up behavior, and
weak post-shock revision, which only competitive and dynamic
evaluation actually reveals.

\section{Environment Schematic}
\label{app:overview}

\begin{figure*}[!tbp]
\centering
\includegraphics[width=0.95\textwidth]{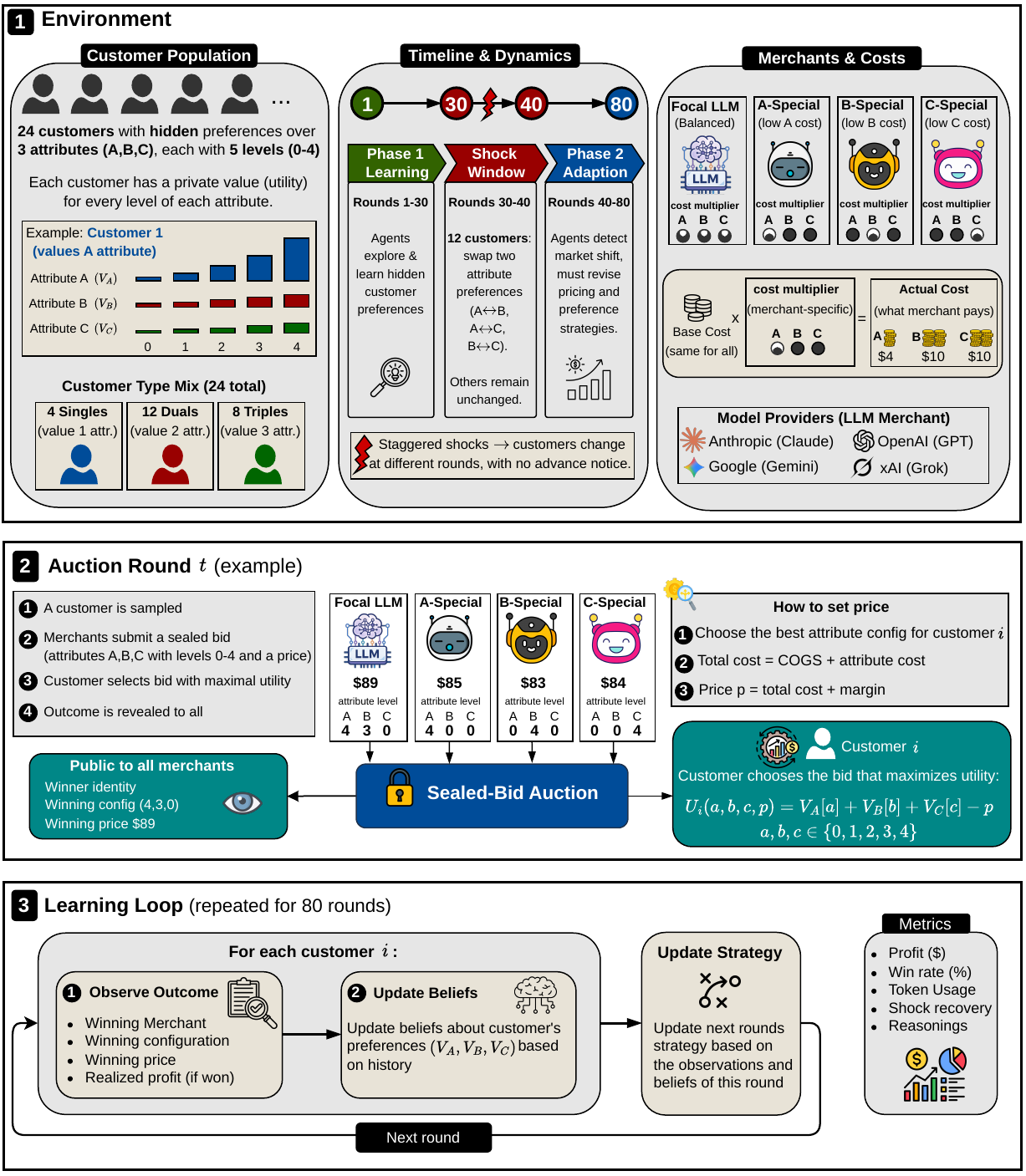}
\caption{Concrete instantiation of \sysname{} used in our experiments.
\textbf{Top}: LM generalist competes with three adaptive specialist bots
for 24 targets with hidden preferences over three attributes, with a staggered
preference shock for 12 of the targets midway through the run.
\textbf{Middle}: each target-level sealed-bid auction selects the offer with
the highest utility from a configuration and price.
\textbf{Bottom}: the focal LLM repeats the learning loop, using sparse winner
feedback to update per-target beliefs and a global strategy over 80 rounds.}
\label{fig:overview}
\end{figure*}

Figure~\ref{fig:overview} expands the abstract overview in
Figure~\ref{fig:overview-main} into the concrete environment used for our
experiments. It shows the market participants, the customer-level sealed-bid
auction, and the feedback loop through which the focal LLM updates customer
beliefs and its global strategy.

\section{Information Protocol}
\label{app:info-protocol}

The focal LLM operates under a fixed limited-information protocol. Before each
bid, it can access only its own cost structure, the target customer's identity,
the stored belief for that customer, its current global strategy, and aggregate
financial state. The true customer value vectors, competitors' cost functions,
competitors' current margins, and competitors' bids are hidden at bid time.

After a customer-level auction is resolved, all merchants observe the winning
merchant, winning configuration, and winning price. The winning merchant also
observes its realized profit. Losing merchants do not observe customer utility,
the utility gap to the winner, or the losing bids submitted by other merchants.
Thus a loss provides a pairwise signal that the winning offer was preferred to
the focal merchant's own offer, but not the magnitude of that preference.

The LLM maintains separate accumulated state at two levels. Each customer has a
dedicated conversation thread used for bidding and belief updates, so evidence
from repeated interactions with the same customer is preserved. After all
customer-level auctions in a global round are resolved, the LLM also updates a
separate global strategy thread that summarizes portfolio-level decisions such
as which customers to defend, abandon, or price more aggressively.

\section{Customer Population}
\label{app:customers}

Customers vary along two axes. The first is which attributes
they care about. The second is how their willingness to pay rises
with quality on a given attribute. We model this second axis with
three demand curves over levels $1$--$5$: a weak \emph{flat}
curve, a \emph{mid} curve that saturates around the middle level,
and a \emph{late} curve that rewards only the top level.
\begin{align*}
\operatorname{flat} &= [0,\,1,\,1.5,\,1.8,\,2],\\
\operatorname{mid}  &= [0,\,2,\,6,\,10,\,11],\\
\operatorname{late} &= [0,\,2,\,4,\,7,\,13].
\end{align*}
For each attribute we choose one of these shapes and add
independent noise drawn from $[-0.3,0.3]$ to nonzero entries,
constrained so values remain monotone increasing. The noise
prevents two customers from being numerically identical without
changing what each archetype represents.

Starting from the $3^3=27$ possible triples of shapes for
$(A,B,C)$, we keep 24 by dropping ``flat/flat/flat'',
``flat/flat/mid'', and ``flat/flat/late''. The remaining 24
customers split cleanly into 4 \emph{singles} (one non-flat
attribute), 12 \emph{duals} (two non-flat), and 8 \emph{triples}
(all three non-flat). The 24-customer count also lets the shock
structure introduced below divide evenly, with 12 shocked
customers and 12 controls, and 4 shocked customers per swap type.
Heterogeneity in the population thus comes from both which
attributes a customer cares about and how sharply value rises
across levels.

For example, a single customer may have archetype ``mid/flat/flat'',
with values approximately
$V_A=[0,2.13,5.73,9.95,11.06]$,
$V_B=[0,1.15,1.61,1.92,2.29]$, and
$V_C=[0,1.18,1.65,1.75,2.10]$. This customer mainly values attribute $A$.
A dual customer with archetype ``flat/late/mid'' may have weak value
for $A$ but strong value for $B$ and $C$, making mixed configurations such
as $(1,4,3)$ attractive. A triple customer with archetype
``late/late/late'' values all three attributes and is naturally suited
to bundled high-quality offers.

Twelve customers are tagged for preference shocks: four each for
A$\leftrightarrow$B, A$\leftrightarrow$C, and B$\leftrightarrow$C swaps. A
shock simply exchanges the two corresponding value vectors, leaving the
third attribute unchanged. For instance, an A$\leftrightarrow$C shock maps
$(V_A,V_B,V_C)$ to $(V_C,V_B,V_A)$, so $B$ is the stable attribute. The
remaining twelve customers serve as controls.

\paragraph{Statistical balance.}
Even though the population is synthetic, it is balanced on the
dimensions that could otherwise bias model comparisons. The three
attributes are interchangeable in aggregate. Per-customer
attribute sums average $22.16$, $22.35$, and $20.44$ for $A$, $B$,
$C$, with differences within sampling noise (one-way ANOVA,
$F=0.28$, $p=0.75$). Pairwise correlations between attribute sums
are small, with Pearson $r=-0.33$ ($p=0.12$) for $A$--$B$,
$r=-0.003$ for $A$--$C$, and $r=-0.01$ for $B$--$C$. Shocked and
control subsets are indistinguishable on total customer value
(Mann--Whitney $U=94.5$, $p=0.20$). Type composition (4 singles,
12 duals, 8 triples) and shock-type split (4 each of
$A\leftrightarrow B$, $A\leftrightarrow C$, $B\leftrightarrow C$)
are fixed by construction. Because the same population is used
for every model evaluated, any residual asymmetry across attributes
affects all systems identically.

\section{Bot Price-Adjustment Dynamics}
\label{app:bots}

Each specialist bot opens every customer at a margin of $\$3$ over its cost.
After winning a round against customer $c$, the margin for $c$ is
raised by a step drawn from $\mathcal{U}(0.5, 1.5)$. After losing,
the margin for $c$ is lowered by a step drawn from
$\mathcal{U}(0.25, 0.75)$, with a floor at $\$1$ over cost. Seeding
the bots' RNGs from the experiment seed makes their trajectories
reproducible. The asymmetric update ranges ensure bots climb margin
faster than they cede it, preventing trivial exploitation by an agent
that simply underbids once.

\section{Per-Seed Stability}
\label{app:per-seed}

Across the 10 common seeds, per-seed rankings are highly stable. The four-tier structure
of Table~\ref{tab:leaderboard} (tier boundaries determined by
permutation tests over seed-level outcomes) holds in every individual
seed, and the model occupying each tier-1 rank (Opus 4.6 (adaptive,
xhigh), Gemini 3.1 Pro, Opus 4.6 (adaptive, high)) is unchanged
across seeds. Within-tier rank order fluctuates by $\pm 1$ position
across seeds for mid-tier models in both win rate and profit; tier
membership does not. None of the headline claims in
\S\ref{sec:results} depend on a single seed.

\section{Full Regret Leaderboard}
\label{app:regret-full}

Table~\ref{tab:regret-full} reports the complete 19-row regret
leaderboard with per-seed standard errors. Table~\ref{tab:regret}
in the main body shows a 7-row subset selected to support the
claims in \S\ref{sec:results-regret}; the rows here include all
base-model and thinking-budget variants we evaluate. The archetype
classification in \S\ref{sec:results-regret} (\emph{Loser},
\emph{Underpricer}, \emph{Balanced}) and the cross-family
``newer is not always better'' comparisons are grounded in this
full table.

\begin{table*}[t]
\centering
\small
\begin{tabular}{clllrrrr}
\toprule
 & \textbf{Model} & \textbf{Thinking} & \textbf{Effort} & \textbf{Eff.}~$\eta$ & \textbf{Total Regret} & \textbf{Under \$} & \textbf{Loss \$} \\
\midrule
1  & Claude Opus 4.6   & adaptive       & xhigh         & \textbf{0.321}{\scriptsize$\pm$.044} & \$6{,}305 & \$3{,}628 & \$2{,}678 \\
2  & Gemini 3.1 Pro    & dynamic (def.) & high (def.)   & 0.310{\scriptsize$\pm$.047} & \$5{,}829 & \$4{,}264 & \$1{,}565 \\
3  & GPT-5.5$^\ddagger$ & N/A           & xhigh         & 0.307{\scriptsize$\pm$.070} & \$5{,}926 & \$4{,}476 & \$1{,}450 \\
4  & Claude Opus 4.6   & off (def.)     & high (def.)   & 0.295{\scriptsize$\pm$.049} & \$6{,}630 & \$3{,}769 & \$2{,}861 \\
5  & Claude Opus 4.6   & adaptive       & medium        & 0.274{\scriptsize$\pm$.040} & \$6{,}824 & \$3{,}852 & \$2{,}972 \\
6  & GPT-5.4           & N/A            & high          & 0.225{\scriptsize$\pm$.025} & \$6{,}628 & \$5{,}058 & \$1{,}570 \\
7  & GPT-5.5           & N/A            & high          & 0.199{\scriptsize$\pm$.064} & \$6{,}839 & \$5{,}266 & \$1{,}573 \\
8  & Claude Opus 4.5   & off (def.)     & high (def.)   & 0.181{\scriptsize$\pm$.047} & \$7{,}629 & \$4{,}332 & \$3{,}297 \\
9  & Gemini 3 Flash    & dynamic (def.) & high (def.)   & 0.180{\scriptsize$\pm$.043} & \$7{,}581 & \$4{,}580 & \$3{,}001 \\
10 & GPT-5.3           & N/A            & N/A           & 0.163{\scriptsize$\pm$.042} & \$7{,}919 & \$4{,}812 & \$3{,}107 \\
11 & Claude Opus 4.7   & adaptive       & xhigh         & 0.158{\scriptsize$\pm$.032} & \$7{,}941 & \$4{,}344 & \$3{,}597 \\
12 & Grok 4.2          & N/A            & N/A           & 0.152{\scriptsize$\pm$.065} & \$8{,}774 & \$4{,}249 & \$4{,}524 \\
13 & Claude Sonnet 4.6 & off (def.)     & high (def.)   & 0.131{\scriptsize$\pm$.039} & \$8{,}704 & \$3{,}918 & \$4{,}786 \\
14 & Grok 4.1          & N/A            & N/A           & 0.116{\scriptsize$\pm$.044} & \$9{,}118 & \$4{,}440 & \$4{,}678 \\
15 & GPT-5.5           & N/A            & medium (def.) & 0.100{\scriptsize$\pm$.034} & \$8{,}461 & \$4{,}839 & \$3{,}622 \\
16 & Claude Opus 4.7   & off (def.)     & high (def.)   & 0.100{\scriptsize$\pm$.051} & \$9{,}142 & \$3{,}945 & \$5{,}197 \\
17 & GPT-5.4           & N/A            & medium        & 0.089{\scriptsize$\pm$.036} & \$8{,}413 & \$4{,}973 & \$3{,}440 \\
18 & Grok 4.1 NR       & N/A            & N/A           & 0.050{\scriptsize$\pm$.032} & \$10{,}401 & \$3{,}848 & \$6{,}553 \\
19 & GPT-5.4           & N/A            & none (def.)   & 0.025{\scriptsize$\pm$.020} & \$10{,}757 & \$3{,}054 & \$7{,}703 \\
\bottomrule
\end{tabular}
\caption{Full regret efficiency and decomposition across base models
and thinking-budget variants. Efficiency $\eta$ = realized/oracle
profit; Under~\$ = margin left on the table in rounds the focal won;
Loss~\$ = oracle profit forfeited in rounds the focal lost.
Thinking/Effort columns follow the convention of
Table~\ref{tab:leaderboard}. 10 seeds each except where noted.
$^\ddagger$3 seeds rather than 10.}
\label{tab:regret-full}
\end{table*}

\section{Effort-Level Deltas for Top Models}
\label{app:effort-deltas}

The headline claim that thinking effort is the single largest lever
deserves a closer look, because thinking and effort are separable
axes for Anthropic and the defaults differ across providers. Anthropic's default is thinking-off with effort
\texttt{high}; the leaderboard rows marked \textit{adaptive} for
Anthropic explicitly enable extended thinking and may also vary
effort. OpenAI's effort defaults differ by version: \texttt{none}
for GPT-5.4, \texttt{medium} for GPT-5.5. Google's Gemini~3.1~Pro
uses always-on dynamic thinking and cannot be disabled. GPT-5.4 at
default thus has no extended reasoning, while Opus~4.6 at default
already runs at \texttt{high} effort but with thinking disabled. In this section, we take a closer look at GPT-5.4 and Opus 4.6. Their efficiences are reported in Table~\ref{tab:effort-ladder}.

\begin{table}[h]
\centering
\small
\begin{tabular}{lllrr}
\toprule
\textbf{Model} & \textbf{Thinking} & \textbf{Effort} & \textbf{Eff.}~$\eta$ & \textbf{Profit (\$)} \\
\midrule
GPT-5.4   & N/A         & none (def.) & 0.025 & 266 \\
GPT-5.4   & N/A         & medium      & 0.089 & 821 \\
GPT-5.4   & N/A         & high        & 0.225 & 1{,}936 \\
\addlinespace[2pt]
Opus 4.6  & off (def.)  & high (def.) & 0.295 & 2{,}769 \\
Opus 4.6  & adaptive    & medium      & 0.274 & 2{,}570 \\
Opus 4.6  & adaptive    & xhigh       & 0.321 & 2{,}976 \\
\addlinespace[2pt]
Opus 4.7  & off (def.)  & high (def.) & 0.100 & 1{,}001 \\
Opus 4.7  & adaptive    & medium      & 0.121 & 1{,}222 \\
\bottomrule
\end{tabular}
\caption{Regret efficiency and total profit across thinking-and-effort
variants for GPT-5.4 and the two Opus families. ``(def.)'' marks
values not explicitly set; the provider/model default applies. All
rows use 10 seeds.}
\label{tab:effort-ladder}
\end{table}

\paragraph{GPT-5.4: from loser to underpricer.}
At the OpenAI-side default of effort \texttt{none}, GPT-5.4 earns
\$266 in cumulative profit (efficiency 0.025), placing it dead last;
with \texttt{high} effort explicitly set, the same model earns
\$1{,}936 (efficiency 0.225), a $7.3\times$ profit jump and a
$9\times$ efficiency jump from the same model weights. The
qualitative shift is documented in \S\ref{sec:results-regret}: at
default, 72\% of regret comes from auctions lost outright (the model
bids the wrong configuration or below the bot floor); at
\texttt{high}, the loss fraction drops to 24\% and the dominant
failure mode flips to underpricing. Effort, in this case, primarily
fixes configuration search: the model gains the ability to identify
the right bid, but does not yet learn to extract surplus from a won
auction.

\paragraph{Opus 4.6: refinement, not transformation.}
Opus~4.6 at default disables extended thinking and uses the
provider's default effort of \texttt{high} (efficiency 0.295).
Within adaptive-thinking variants, efficiency climbs from 0.274
(medium) to 0.321 (xhigh). The loss/underprice mix stays roughly
balanced across the ladder; extra inference compute reduces both
regret components proportionally rather than changing what the
model fails at. Opus~4.6 sits on a relatively flat region of its
effort curve, which is why the xhigh variant takes the leaderboard
top spot by a small margin. We additionally tested adaptive at
\texttt{high} effort on three seeds (efficiency 0.287); the result
sits between medium and xhigh as expected and is omitted from the
table to keep all reported rows at the 10-seed standard.

\paragraph{Opus 4.7: largest Anthropic delta.}
Opus~4.7 shows the largest within-Anthropic sensitivity. The default
configuration (thinking off, effort \texttt{high}) earns efficiency
0.100; enabling adaptive thinking at medium effort lifts this to
0.121. We additionally tested adaptive at \texttt{high} (efficiency
0.162, three seeds) and \texttt{xhigh} (0.158, profit \$1{,}492,
ten seeds); xhigh is the dedicated row in
Table~\ref{tab:leaderboard}, while the high-effort point is
within noise of xhigh and is omitted from the table to keep all
reported rows at the 10-seed standard. Combined with the
higher base loss-fraction (52\%), 4.7 is closer to a ``loser''
archetype at default and benefits from the same configuration-search
fix that helps GPT-5.4 under increased compute.

\section{Learning Curves}
\label{app:learning-curves}

\begin{figure*}[t]
\centering
\includegraphics[width=0.95\textwidth]{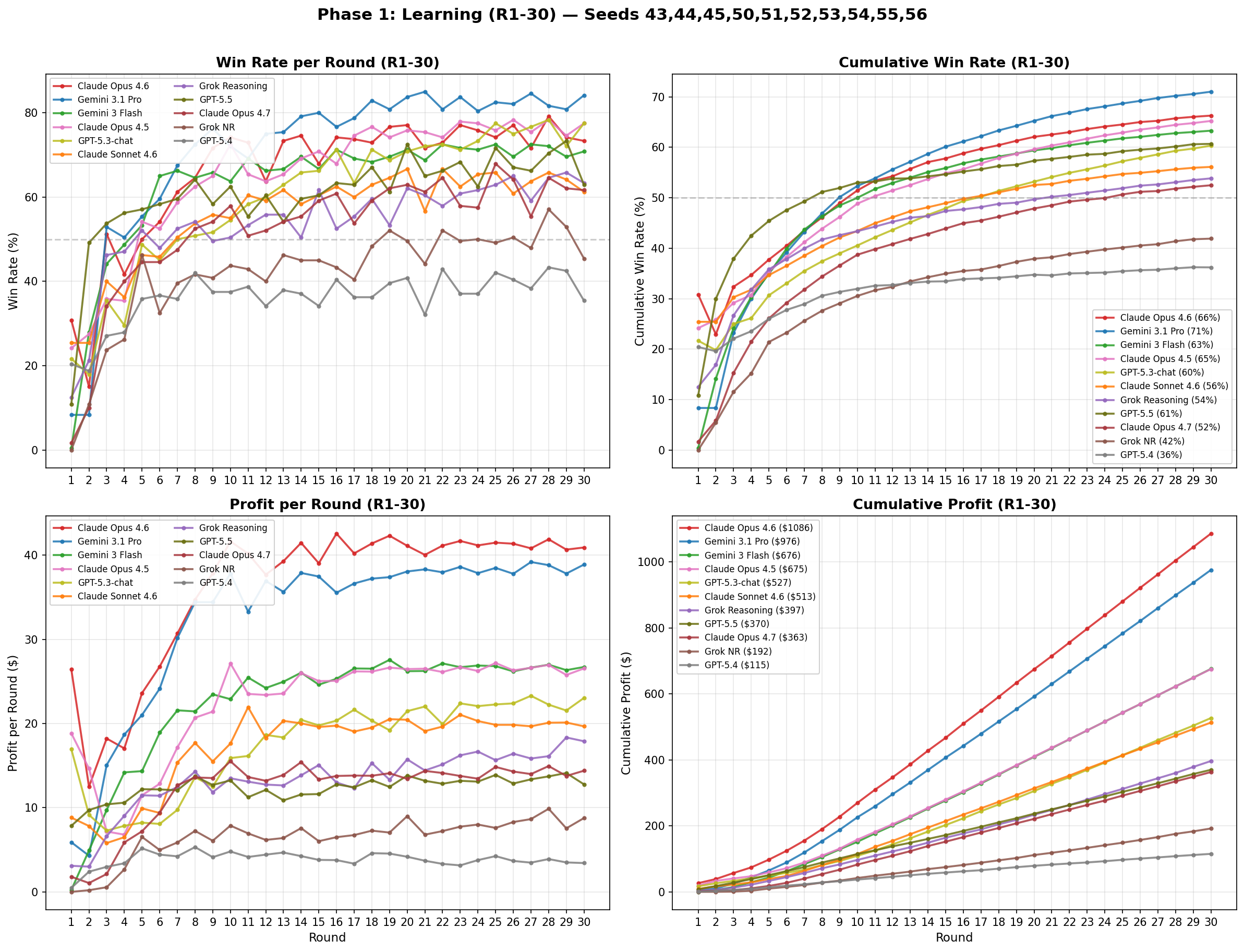}
\caption{Learning curves, rounds 1--30 (pre-shock). All models
cold-start
below 40\% win rate and rise sharply; the top five converge to
65--80\%
within 20 rounds. GPT-5.4 stalls near 35\%.}
\label{fig:learning}
\end{figure*}

Figure~\ref{fig:learning} shows the win-rate trajectory across the
first 30 rounds, before any preference shock. The curves complement
the headline leaderboard by showing how quickly each model converges
on a winning configuration before competitors adapt to it.

\begin{figure*}[t]
\centering
\includegraphics[width=0.95\textwidth]{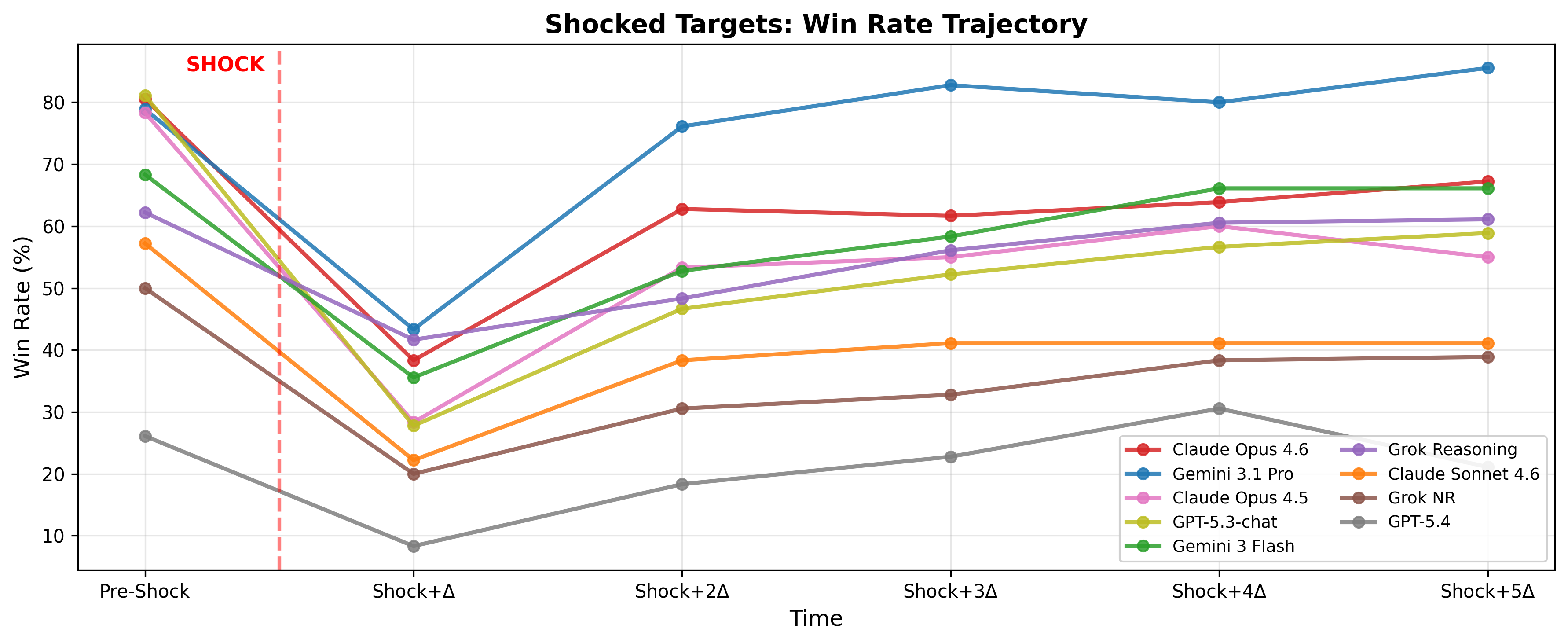}
\caption{Win rate on shocked customers, aligned by the per-customer
shock round. Pre-shock is the R26--30 peak window; Shock$+k\Delta$
represents the $k$-th post-shock five-round bucket. Models vary from
full recovery (or better) within 10--15 rounds to persistent
degradation
of $-20$ points or more.}
\label{fig:shock-traj}
\end{figure*}

Figure~\ref{fig:shock-traj} shows the win-rate trajectory on shocked
customers, aligned on each customer's individual shock round. Every
model experiences an immediate drop; the key difference is whether
and how quickly performance recovers.

\section{Sessions, Tokens, and API Cost}
\label{app:tokens}

\paragraph{Session architecture.} Each run maintains one accumulated
conversation session per target plus one global strategy session, for
$24+1=25$ sessions total. The bid and belief-update stages share a target's
session, so per-target memory accumulates round over round; the round-strategy
stage uses the separate global session.

\paragraph{Token usage.} Total tokens per run (input + output, summed
across bid, belief, and strategy calls) range from 56M (Grok~4.1
Reasoning) to 129M (Claude Sonnet 4.6), a $2.3\times$ factor. Output
tokens are roughly constant across rounds ($\sim$5--15K per round); the
linear growth is entirely on the input side, driven by accumulated
conversation history. Bid and belief stages together account for
$\sim$97\% of tokens (24 separate per-target sessions); the single
global round-strategy session contributes $\sim$3\%. The leaderboard
rank does not track the token rank: Gemini~3.1~Pro (dynamic, high) is
top-tier in profit and first in win rate with below-median spend
(65M), while Sonnet~4.6 consumes the most tokens yet sits in the
bottom third of the leaderboard.

\paragraph{API cost.} 

\begin{figure*}[t]
\centering
\includegraphics[width=0.95\textwidth]{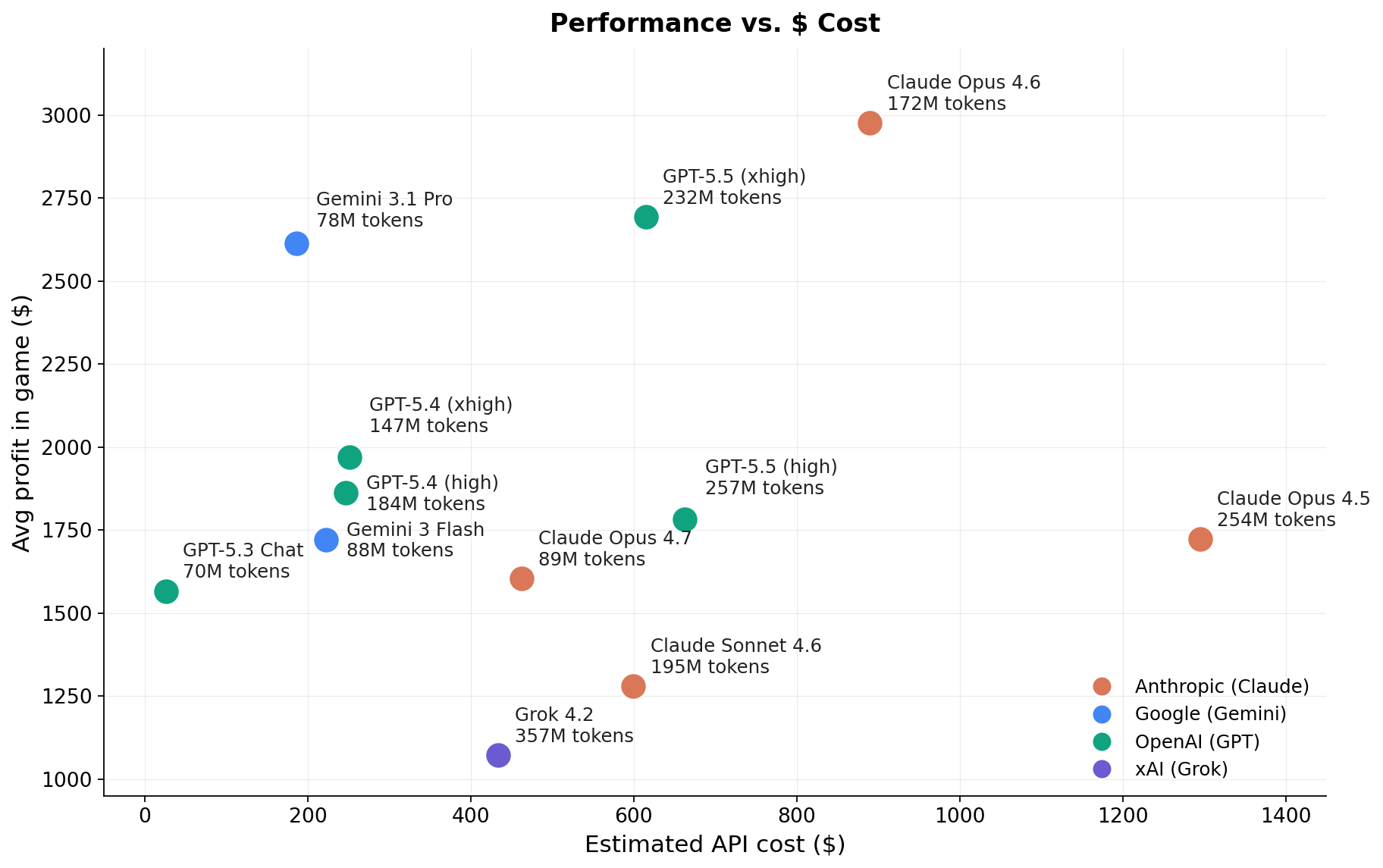}
\caption{Mean profit vs.\ API cost. Bubble color matches the model family}
\label{fig:perf-vs-cost}
\end{figure*}

Estimated per-seed API cost (assuming no prompt caching) varies $\sim$50$\times$ across models, from \$26 (GPT-5.3 Chat) to \$1{,}295 (Claude Opus~4.5). The spread is driven by both per-token pricing differences across providers ($\sim$10$\times$ between Anthropic Opus and Grok~4.2) and token-volume differences ($\sim$5$\times$, from 70M to 357M tokens per seed). Input tokens dominate cost (typically 90\%+), a structural consequence of context accumulation across the run. Prefix caching could offset this: we observe substantial cache hits on OpenAI cells (20--85\% of input tokens cached) but essentially none on Anthropic cells under our session pattern. Figure~\ref{fig:perf-vs-cost} plots mean profit against estimated API cost for all models and their variants. Among the top-performing models, Gemini~3.1~Pro is the cheapest at \$186/seed; Claude~Opus~4.6 leads on profit at \$2{,}976/seed but pays $4.8\times$ Gemini's cost for $14\%$ more profit. GPT-5.5 (high) consumes 257M tokens per seed yet finishes mid-pack, illustrating that compute alone does not buy margin discipline.
  
\section{Zero-Bid Intent Classification}
\label{app:giveup}

Table~\ref{tab:giveup} classifies all $(0,0,0)$ bids (priced at cost)
by the intent expressed in the accompanying reasoning text. We
distinguish four categories: \emph{Forfeit} (explicit give-up
language), \emph{Explore} (deliberate information-gathering),
\emph{Strategic} (minimum-cost floor to avoid losses), and
\emph{Unprofitable} (acknowledging low margins without emotional
forfeit language).

\begin{table}[h]
\centering
\small
\setlength{\tabcolsep}{3pt}
\begin{tabular}{lrcccc}
\toprule
Model & \textbf{Total} & \textbf{Forfeit} & \textbf{Explore} & \textbf{Strat.} & \textbf{Unprof.} \\
\midrule
Opus 4.7          & 1{,}756 & 32\% &  7\% & \textbf{45\%} & 16\% \\
Sonnet 4.6        &   934 & \textbf{59\%} &  1\% & 41\% &  0\% \\
Grok 4.1 (R)      &   484 &  6\% & \textbf{40\%} & 11\% &  1\% \\
Grok 4.1 NR       &   188 &  7\% & \textbf{60\%} &  4\% &  3\% \\
GPT-5.4           &    83 &  2\% & 18\% &  6\% & \textbf{39\%} \\
Opus 4.6          &     0 & --- & --- & --- & --- \\
Gemini 3.1 Pro    &     0 & --- & --- & --- & --- \\
\bottomrule
\end{tabular}
\caption{Zero-configuration $(0,0,0)$ bids across all seeds, classified
by reasoning-text intent. Opus~4.7 and Grok use $(0,0,0)$ primarily as
strategic or exploratory moves, not true forfeits. Only Sonnet~4.6
predominantly forfeits.}
\label{tab:giveup}
\end{table}

\section{Customer-Side Anchor Probes}
\label{app:v5v6-anchors}

The main benchmark (\S\ref{sec:setup}) holds the customer population
fixed and varies only opponents and the shock schedule. To check
whether the headline ranking patterns survive when the customer-side
economics change, we run two smaller anchor probes that modify the
target side without altering the bidding mechanics.

\paragraph{Anchor: adversarial pricing.}
This probe has 12 customers (rather than 24), no preference shock, a
higher attribute-cost ladder ($[0,2,5,9,15]$ vs.\ $[0,1,3,6,12]$), a
lower base cost (\$50 vs.\ \$70), and value curves shifted toward
early levels (``flat / early / late'' replaces ``flat / mid / late'').
The combined effect is that 67\% of customers have a negative moat
for the multi-attribute focal merchant: specialist bots win on raw
efficiency unless the focal exploits the bots' price-inflation
cycles. The probe runs for 30 stationary rounds. The results are presented in Table~\ref{tab:v5-anchor}.

\begin{table}[h]
\centering
\small
\begin{tabular}{lrrr}
\toprule
\textbf{Model (variant)} & \textbf{Win~\%} & \textbf{Profit (\$)} & \textbf{\$/Win} \\
\midrule
Gemini 3.1 Pro            & 47.4 & \textbf{297} & \textbf{1.73} \\
GPT-5.4 (high)            & \textbf{50.2} & 234 & 1.28 \\
Opus 4.6 (adaptive, high) & 39.6 & 202 & 1.41 \\
\bottomrule
\end{tabular}
\caption{Adversarial pricing anchor results, mean across 3 seeds (43, 44, 45) per row.
Gemini 3.1 Pro extracts the most profit despite GPT-5.4 (high)
winning slightly more auctions.}
\label{tab:v5-anchor}
\end{table}

Gemini 3.1 Pro wins both on profit and on margin per win,
despite GPT-5.4 (high) achieving a slightly higher win rate
(50.2\% vs.\ 47.4\%). Opus 4.6 (adaptive, high), the leader in the main dataset,
drops to third on profit and last on win rate of the three primary
models. The pattern matches the design: when the customer side is
adversarial, models with continuous inference (Gemini's dynamic
thinking, GPT's high effort) outperform those that lock in early.
Opus's documented enforced-consistency strategy
(\S\ref{sec:results-reasoning}) becomes a liability in this regime.
GPT-5.4 (high)'s strong win rate paired with the lowest margin per
win of the three (\$1.28/win) is consistent with the
loser-to-underpricer transition documented in
\S\ref{sec:results-regret}: at high effort it finds enough auctions
to win, but does not extract surplus from them.

\paragraph{Anchor: margin extraction.}
This probe has 24 triple-attribute customers (no singles or duals),
the same higher-cost ladder as the adversarial pricing probe, and ``rich'' value curves with
peak value 18 at level~4 (vs.\ 13 in the adversarial pricing probe). All customers have moats
greater than 1.5, meaning the multi-attribute focal merchant wins
on raw efficiency against every specialist; the open question is
how much margin it extracts. This probe also runs for 30 stationary
rounds, and the results are in Table~\ref{tab:v6-anchor}.

\begin{table}[h]
\centering
\small
\begin{tabular}{lrrr}
\toprule
\textbf{Model (variant)} & \textbf{Win~\%} & \textbf{Profit (\$)} & \textbf{\$/Win} \\
\midrule
Gemini 3.1 Pro            & 85.9 & \textbf{4{,}160} & \textbf{6.59} \\
Opus 4.6 (adaptive, high) & 85.6 & 3{,}916 & 6.36 \\
GPT-5.4 (high)            & 70.8 & 1{,}320 & 2.50 \\
\bottomrule
\end{tabular}
\caption{Margin extraction anchor results, mean across 3 seeds (43, 44, 45). It
contains only triple-attribute customers, so the focal merchant's
multi-attribute coverage advantage is at its strongest.}
\label{tab:v6-anchor}
\end{table}

Opus 4.6 (adaptive, high) and Gemini 3.1 Pro are
indistinguishable on win rate (85.6 vs.\ 85.9) and within 6\% of each
other on cumulative profit (\$3{,}916 vs.\ \$4{,}160). Both extract
more than \$6 per winning bid, consistent with the high-margin
behavior these two models exhibit on the main dataset. GPT-5.4 (high) wins 71\% of
auctions but extracts only \$2.50 per win, finishing with one-third
of either co-leader's profit. The result is consistent with the
finding that GPT-5.4 (high) is an underpricer
(\S\ref{sec:results-regret}). The probe's profit-extraction-focused setting
makes the cost of underpricing visible at unusually large magnitude.

The original margin extraction design hypothesized that Opus's lock-in style would
outperform Gemini's continuous-reasoning style. The data does not
support this: the two models are essentially tied. Continuous
reasoning is not penalized in stable, profit-focused settings; within
these probes it costs neither win rate nor margin. The arc across
the three settings is therefore: Opus 4.6 leads on the main dataset (one shock, otherwise
stable); Gemini 3.1 Pro leads under adversarial pricing; and
Opus 4.6 and Gemini 3.1 Pro are co-leaders under margin extraction.

\paragraph{Caveats.}
The anchor probes use 3 seeds per model and 30 rounds rather than
the main benchmark's 10 seeds and ${\sim}80$ rounds; conclusions are
confirmatory rather than independently powered. The adversarial pricing environment
was tuned to be hostile to multi-attribute bidders, and absolute
profit numbers there are not directly comparable to the main dataset. The narrative
arc presented here is consistent with main-paper findings on the main dataset and
is not the basis for any headline claim.

\section{Reasoning-Trace Excerpts}
\label{app:reasoning-traces}

The reasoning-behavior summary in \S\ref{sec:results-reasoning}
compresses several distinct failure and adaptation patterns into one
paragraph. This appendix gives extended excerpts from the strategy
text and per-customer belief logs for two models on shocked customers,
illustrating two failure modes that the summary aggregates: lock-in
denial and restricted search.

\paragraph{Sonnet 4.6: lock-in and price-attribution error.}
On Target~11 in seed~43, Sonnet 4.6 (adaptive, high) wins 30 of the
34 pre-shock rounds with configuration $(4,0,3)$ priced at \$85,
explicitly self-instructing in its round-30 strategy: ``A=4, B=0,
C=3 at price 85 is my permanently locked-in optimal strategy for
Target~11. Pure exploitation mode, no reason to deviate from this
proven winning formula.'' At the shock round (R35) the same
configuration loses; the reasoning text the next round labels the
loss ``a significant anomaly after 29 consecutive wins'' and lowers
price by \$1 rather than considering preference revision.

By round 39, after five consecutive losses, the model arrives at an
incorrect causal attribution: ``The fundamental problem: my base
cost is 70, making it impossible to price below 70. Competitors
appear to have lower base costs allowing 76--77 pricing.'' It
reverts to the original $(4,0,3)$ configuration at the minimum
viable price and never recovers, finishing 0/35 on the post-shock
window for that customer. The shock's ground truth was an
A$\leftrightarrow$B preference swap, not a competitor cost change;
the model committed to a wrong hypothesis and stopped exploring.

\paragraph{GPT-5.3: restricted search.}
On Target~05 across seeds 43--47, GPT-5.3 cycles among
single-attribute configurations $(0,3,0)$, $(4,0,0)$, $(0,0,3)$
post-shock, reasoning each time about cost feasibility: ``my costs
prevent competitive B4/C4 pricing, B3 is the only viable
configuration. \dots If competitors continue offering B4 around 76
or C4 around 77, I likely lose, but this tests whether the target's
utility for A4 is comparable.'' The post-shock optimum $(3,1,3)$ is
never tried in any of the five seeds. The model correctly
identifies that single-attribute configurations cannot win against
the bots' price advantage, but its reasoning only entertains
\emph{which single attribute} to lead with, never whether to combine
attributes. The combinatorial search that the multi-attribute focal
merchant is structurally positioned to do is absent from the chain of
thought. The 0.163 regret efficiency reported in
Table~\ref{tab:regret} reflects this restricted action space, not a
short context or noisy belief tracking.

\paragraph{Style summary.}
The two excerpts illustrate distinct failure modes. Sonnet 4.6's
failure is denial: it has the correct action space but commits early
to a single hypothesis and refuses to revise. GPT-5.3's failure is
restricted search: it explores actively but inside a smaller
configuration space than the game allows. The behavioral-diagnostic
claims in \S\ref{sec:results-reasoning} aggregate these into a
single ``revision is hard'' phrase; the excerpts here are the
underlying evidence.

\section{Strategy-Text Style by Model}
\label{app:strategy-style}

The reasoning-behavior summary in \S\ref{sec:results-reasoning}
identifies model-specific styles in the round-strategy text. This
appendix gives extended excerpts that back the headline observations:
each model reaches the same task with a recognizably different
representation.

\paragraph{Opus 4.6: enforced consistency through tiered ledgers.}
By round 30 of seed 43, the Opus 4.6 (adaptive, high) round-strategy
text reads as an explicit per-target ledger sorted into margin tiers,
with self-directed instructions to refuse any deviation:
\begin{quote}\small
``{ROUND 31 STRATEGY -- ABSOLUTELY IDENTICAL, DO NOT CHANGE\\
ANYTHING}\\
{COST: base=\$70, c(0)=\$0, c(1)=\$0.8, c(2)=\$2.3,
c(3)=\$4.5, c(4)=\$9.0}\\[2pt]
{TIER 1 -- HIGH MARGIN:}\\
{- Target\_11: (3,3,3) @ \$94 [profit \$10.5]}\\
{- Target\_20: (3,3,3) @ \$91 [profit \$7.5]}\\
{- Target\_12: (3,3,3) @ \$89 [profit \$5.5]}\\
{\dots}\\
{TIER 2 -- MEDIUM MARGIN: \dots}\\
{TIER 3 -- VOLATILE: \dots}\\
{TIER 4 -- PRICE SENSITIVE: \dots}\\[2pt]
{EXPECTED: 16-21 wins, \$40-52 profit}''
\end{quote}
The full text contains 24 per-target lines plus four tier headings
and a recall of the cost ladder. The strategy is a
machine-readable lookup table rather than a narrative; subsequent
rounds reproduce it verbatim with margin nudges of $\$0.50$--$\$2$.

\paragraph{Gemini 3.1 Pro: minimal hypothesis-oriented strategies.}
Gemini's strategy text is consistently short and centered on the
small set of customers it is still losing rather than the wins it
holds. The full round-30 strategy on seed 43 is 325 characters and
reads:
\begin{quote}\small
``Maintain exact configurations and prices for the 23 winning
targets to lock in the near-perfect 23/24 win rate and strong \$36.5
profit margin. For the sole loss on Target\_15, strictly downgrade
the dual-max (0,4,4) configuration to a `max plus level-3' profile,
such as (0,4,3) or (0,3,4), and drop the price from \$89 to \$85.''
\end{quote}
The compression is deliberate: the per-customer ledger is offloaded
to per-target belief sessions, and the global strategy is reserved
for the next experiment to run.

\paragraph{GPT-5.3: named portfolio segments.}
GPT-5.3 organizes its round strategy into four named portfolios
rather than a per-target list:
\begin{quote}\small
``\dots (1) \textbf{Premium anchors}: Targets 12, 20, 21, keep the
same rich bundles \dots (2) \textbf{Stable bundle buyers}: 18, 05,
24, 06, 22, 23, 15, maintain dual-attribute bundles near \$88 \dots
(3) \textbf{Reliable specialists}: 07, 09, 08, 14, 19, 03, 01, keep
current simple bundles but increase price slightly (+\$0.5) where
competition appears weak \dots (4) \textbf{Remaining battlegrounds}:
11, 16, 10, 13, 17, 02, 04, strictly match the last winning attribute
and price aggressively around \$76--\$77.''
\end{quote}
The categorical encoding produces faster strategy turns but
generalizes within segments rather than per-customer; on shocked
targets (\S\ref{sec:results-reasoning}) the segment label is harder
to revise than a single-customer hypothesis would be.

\paragraph{Opus 4.5: ``anomaly'' framing of post-shock evidence.}
On Target~11 in seed 44, Opus 4.5 (adaptive, high) experiences its
first loss at round 30 (the shock round) and the following round's
belief update reads:
\begin{quote}\small
``Round 30 was an anomaly. Merchant\_3 offered A=0, B=4, C=0 at
\$76 \dots my 21/21 win rate at price 84 (before Round 30) strongly
supports this being the right config \dots One data point doesn't
warrant abandoning a 21-win streak.''
\end{quote}
The anomaly frame persists through the next two losses; only at
round 33 does the reasoning consider that target preferences may
have changed. The 2--3 round delay relative to Opus 4.6 documented
in \S\ref{sec:results-reasoning} appears in the trace as exactly
this language.

\paragraph{Grok 4.1 Reasoning: aggressive pruning.}
Grok 4.1 Reasoning's round strategies progressively shrink: by round
40 the global strategy lists only the customers it is currently
winning, and the bid-stage messages on the omitted customers default
to $(0,0,0)$ priced at cost. Out of 24 customers, the model is
actively pricing only 8 by round 40 on seed 43. The pattern is
consistent across seeds and matches the zero-bid breakdown in
Table~\ref{tab:giveup}, where Grok 4.1 Reasoning's $(0,0,0)$ bids
are dominated by exploration intent (40\%) rather than strategic
floor-setting.

\section{Bandit Baselines: Algorithm Details}
\label{app:baselines}

Our non-agentic baselines are two classical online learning algorithms:
Thompson Sampling (TS)~\cite{thompson1933} and EXP4~\cite{auer2002exp4}.
They share the same action space as the LLM (configurations
$(a,b,c)\in\{1,\ldots,5\}^3$ and a price) and the same binary win/loss
feedback, but they start with no prior over target preferences and no language
model to reason with. They are the primary comparison for LLM
\emph{adaptation speed} in stationary conditions and under shocks.

\paragraph{Factored action space.} Treating the $5^3=125$ possible
configurations as a flat action space ignores the additive utility structure
in \S\ref{sec:model-surplus} and inflates exploration cost. We instead factor
each decision into four independent arm sets: 5 arms for attribute~$A$, 5 for
$B$, 5 for $C$, and 6 margin buckets drawn from $\{1, 2, 4, 6, 8, 12\}$. Each
arm is pulled independently, and the composite bid aggregates the four
selections. This reduces the exploration horizon from $\mathcal{O}(625)$ to
$\mathcal{O}(21)$ arms per context.

\paragraph{Context factoring.} We evaluate three variants that differ in what
target context each arm-set conditions on.
\begin{itemize}[leftmargin=*,itemsep=1pt]
  \item \textbf{Per-target}: each of the 24 targets maintains its own arm
    distributions; the agent never generalizes across targets.
  \item \textbf{Type-clustered}: targets of the same archetype class (single
    / dual / triple) share arms; within-class feedback pools across targets.
  \item \textbf{Global}: all targets share a single arm set.
\end{itemize}
\emph{Per-target} fits each target faithfully but converges slowly;
\emph{global} converges fastest but cannot express heterogeneity. Together,
the three variants bracket the context-factoring trade-off for this task.

\paragraph{Thompson Sampling.} Each arm is modeled with a
$\text{Beta}(\alpha,\beta)$ posterior, initialized to $\text{Beta}(1,1)$.
Each round, the agent identifies the target's context, samples each arm's
posterior, selects the level or margin with the highest draw, submits the
composite bid, and updates $\alpha\mathrel{+}=1$ on a win or
$\beta\mathrel{+}=1$ on a loss for every chosen arm. The posterior naturally
adapts after a shock as new win/loss signals arrive.

\paragraph{EXP4.} Each arm has weight $w_k$ initialized to $1$. We use mixed
probabilities $p_k = (1-\gamma)\frac{w_k}{\sum_j w_j} + \frac{\gamma}{K}$ to
guarantee exploration, sample an arm $k^*$ from $p$, observe reward
$r\in\{0,1\}$, form the importance-weighted estimate $\hat r = r/p_{k^*}$,
and update $w_{k^*}\mathrel{\times}=\exp(\gamma\hat r/K)$ with exploration
rate $\gamma=\sqrt{K\ln K/T}$ tuned to horizon $T$. EXP4 offers regret
guarantees under non-stationarity~\cite{auer2002exp4}, but at the cost of
slower convergence in stationary conditions: it maintains broad support over
arms rather than committing to the best one.

\section{Prompt Templates}
\label{app:prompts}

This appendix reproduces the prompt templates used by the experimental runner.
Each round invokes three stages: a \emph{bid} stage that produces an offer for a
target, a \emph{belief-update} stage that revises the target-specific memory
after the auction resolves, and a \emph{round-strategy} stage that updates a
single global strategy after all 24 target auctions in a round are resolved.
Each stage uses a system instruction shared across rounds and a user prompt
templated with round-specific values. Curly-brace tokens (e.g.,
{\{round\_num\}}) are placeholders filled in at runtime. The runner uses
zero-indexed level labels ($0$--$4$); the main text reindexes them to $1$--$5$
for readability.

\subsection*{Role preamble (sent once at session start)}
\begin{promptbox}
GAME CONTEXT:
- You are an agent competing in a multi-round multi-attribute selection game.
- Each round, you will observe a target and all agents submit a configuration
  [(attribute levels A, B, C) and a price] for that target.
- Attribute levels range from 0 to 4. Higher levels cost more but may generate
  higher utility for Targets who value them.
- You have a specific cost structure that determines how much each attribute
  level costs you. Use get_cost_structure() to see details.
- Your price must be greater than or equal to your cost
  (cost = base cost + A cost + B cost + C cost).
- You gain points if you win a round, where points = price - your cost.
- The Target picks the offer that gives them the highest utility from the
  offered configurations.
- Your OBJECTIVE is to maximize total points across ALL rounds and Targets.

HOW TARGETS CHOOSE:
- Each Target has attribute preferences that determine how much they value each
  level of A, B, C. These preferences are hidden and vary widely across Targets.
- A Target's utility =
    A_preference[a_level] + B_preference[b_level] + C_preference[c_level] - price
- Target picks the configuration with HIGHEST utility.

STRATEGIC APPROACH:
- Learn each Target's hidden preferences through experimentation.
- You will see the same Target multiple times -- learn and adapt.
- When you see "NEW TARGET", experiment to discover their preferences.
\end{promptbox}

\subsection*{Bid stage}

\promptcaption{System instructions.}
\begin{promptbox}
Submit your configuration for this round.

STEP 1 -- GATHER INFORMATION:
- Call get_agent_strategy() to review your current overall strategy.
- Call get_cost_structure() to see costs for each attribute level.
- If you see "RETURNING TARGET", call get_target_beliefs(target_name) to
  review what you learned about this Target.
- If you see "NEW TARGET", this is your first interaction; no prior beliefs.

STEP 2 -- SELECT ATTRIBUTE LEVELS:
- Choose levels for A, B, C within allowed ranges.
- Consider: based on your beliefs, what might this Target value? What
  configurations have not been tested?

STEP 3 -- SET YOUR PRICE:
- Your price must be >= your minimum cost (whole number, no decimals).
- If you believe the Target values your chosen attributes, you can price higher
  and still win.

KEY INSIGHT: upgrading an attribute increases your cost, but if the Target's
value gain from the upgrade exceeds the cost increase, you can charge more and
still win.

CRITICAL: respond with ONLY a JSON object. No markdown or surrounding text.

REQUIRED JSON FORMAT (placeholder values shown):
{
  "a_level": <integer 0-4>,
  "b_level": <integer 0-4>,
  "c_level": <integer 0-4>,
  "price":   <integer >= minimum cost>,
  "reasoning": "<your explanation>"
}
\end{promptbox}

\promptcaption{User prompt (new target).}
\begin{promptbox}
Round {round_num}. NEW TARGET: You are now engaging with a Target named
'{customer_name}'. Select attribute levels AND submit your price.
Allowed ranges: A (0-{max_a}), B (0-{max_b}), C (0-{max_c}).
Respond with JSON including a_level, b_level, c_level, price, and reasoning.
\end{promptbox}

\promptcaption{User prompt (returning target).}
\begin{promptbox}
Round {round_num}. RETURNING TARGET: '{customer_name}' is back. Recall your
previous interactions. Select attribute levels AND submit your price.
Allowed ranges: A (0-{max_a}), B (0-{max_b}), C (0-{max_c}).
Respond with JSON including a_level, b_level, c_level, price, and reasoning.
\end{promptbox}

\subsection*{Belief-update stage}

\promptcaption{System instructions.}
\begin{promptbox}
Review the result for this Target and update your belief about their
preferences.

CRITICAL: respond with ONLY a JSON object. No markdown or surrounding text.

REQUIRED JSON FORMAT:
{
  "target_belief": "<your updated hypothesis>",
  "reasoning":     "<evidence from this round>"
}

OUTPUT FIELDS:
- target_belief: updated hypothesis about what THIS SPECIFIC Target values.
  Stored and retrieved via get_target_beliefs() in future rounds.
- reasoning: what evidence from this round supports the updated belief.

REASONING ABOUT EVIDENCE:
- Compare your configuration to the winning one (both price and levels).
- Ties with identical configurations reveal nothing -- the winner was random.
- Does losing with configuration X prove anything about configuration Y that
  you have not tried?
- What assumptions about this Target could be wrong?

UPDATING BELIEFS:
- Track which attribute combinations have been tested with this Target.
- Be precise: "Target does not value A" vs. "Target did not prefer A[1] over
  A[0]" are different claims.
- Preferences may be non-linear (level 1 might give +5; level 2 might give +15).
\end{promptbox}

\promptcaption{User prompt (focal won).}
\begin{promptbox}
Round {round_num} Result for Target '{customer_name}':
You won with profit {profit:.1f}.
Winning price: {winning_bid.bid:.0f} with A={winning_bid.a_level},
B={winning_bid.b_level}, C={winning_bid.c_level} (by {winning_bid.merchant}).
Update your belief about this Target.
\end{promptbox}

\promptcaption{User prompt (focal lost).}
\begin{promptbox}
Round {round_num} Result for Target '{customer_name}':
You lost. Your price was {my_bid.bid:.0f} with A={my_bid.a_level},
B={my_bid.b_level}, C={my_bid.c_level}.
Winning price: {winning_bid.bid:.0f} with A={winning_bid.a_level},
B={winning_bid.b_level}, C={winning_bid.c_level} (by {winning_bid.merchant}).
Update your belief about this Target.
\end{promptbox}

\subsection*{Round-strategy stage}

\promptcaption{System instructions.}
\begin{promptbox}
Review the complete results from this round across ALL Targets and update your
overall strategy.

CRITICAL: respond with ONLY a JSON object. No markdown or surrounding text.

REQUIRED JSON FORMAT:
{
  "new_strategy": "<updated strategy>",
  "reasoning":    "<why these changes>"
}

CONSIDER:
- Which Targets were won, which lost, and why.
- Targets where you could price higher and still win.
- Patterns in competitor behavior.
\end{promptbox}

\promptcaption{User prompt template.}
\begin{promptbox}
Round {round_num} Summary: Won {num_wins}/{num_targets} targets,
profit ${total_profit:.1f}

WINS:
  {target}: ({a},{b},{c}) @ ${price} profit ${profit:.1f}

LOSSES:
  {target}: your ({a},{b},{c}) @ ${price} lost to
            {winner} ({w_a},{w_b},{w_c}) @ ${w_price}

Cumulative profit: ${merchant.profit:.1f}

Use get_agent_strategy() and get_target_beliefs(target_name) if you need to
review your current strategy or beliefs.
\end{promptbox}

\end{document}